\documentclass[sigconf, screen, nonacm]{acmart}
\usepackage{multirow}
\usepackage{listings} 
\usepackage{xcolor}
\AtBeginDocument{%
  }

\usepackage{subcaption}
\usepackage{makecell}
\usepackage{tabularx}
\usepackage{appendix}
\usepackage{colortbl}
\setcopyright{none} 
\renewcommand\footnotetextcopyrightpermission[1]{}
\begin{document}

\title{S3T-Former: A Purely Spike-Driven State-Space Topology Transformer for Skeleton Action Recognition}



\author{Naichuan Zheng}
\email{2022110134zhengnaichuan@bupt.edu.cn}
\affiliation{%
  \department{School of Information and Communication Engineering}
  \institution{Beijing University of Posts and Telecommunications}
  \city{Beijing}
  \country{China}
}

\author{Hailun Xia}
\authornote{Corresponding author.} 
\email{xiahailun@bupt.edu.cn}
\affiliation{%
  \department{School of Information and Communication Engineering}
  \institution{Beijing University of Posts and Telecommunications}
  \city{Beijing}
  \country{China}
}

\author{Zepeng Sun}
\email{szp2025140107@bupt.edu.cn}
\affiliation{%
  \department{School of Information and Communication Engineering}
  \institution{Beijing University of Posts and Telecommunications}
  \city{Beijing}
  \country{China}
}

\author{Weiyi Li}
\email{liweiyi@bupt.edu.cn}
\affiliation{%
  \department{International School}
  \institution{Beijing University of Posts and Telecommunications}
  \city{Beijing}
  \country{China}
}

\author{Yujia Wang}
\email{wangyujia@bupt.edu.cn}
\affiliation{%
  \department{School of Economics and Management}
  \institution{Beijing University of Posts and Telecommunications}
  \city{Beijing}
  \country{China}
}
\thanks{Code: \url{https://github.com/zhengnaichuan2022/S3T-Former}}
\renewcommand{\shortauthors}{Zheng et al.}
\begin{abstract}
Skeleton-based action recognition is crucial for multimedia applications but heavily relies on power-hungry Artificial Neural Networks (ANNs) built upon dense multiply-accumulate (MAC) operations, limiting their deployment on resource-constrained edge devices. Spiking Neural Networks (SNNs) provide an energy-efficient alternative; however, existing spiking models for skeleton data struggle to comprehensively perceive complex topological motion dynamics. They often compromise the intrinsic sparsity of SNNs by resorting to dense matrix aggregations or non-sparse spectral transformations, while simultaneously suffering from the severe short-term amnesia of spiking neurons. In this paper, we propose the Spiking State-Space Topology Transformer (S3T-Former), the first purely spike-driven architecture specifically designed to resolve these bottlenecks. To explicitly capture multi-order biomechanical dynamics, we formulate the Multi-Stream Anatomical Spiking Embedding (M-ASE) as a generalized kinematic differential operator, elegantly translating structural and temporal variations into rich event streams. Synergistically, an Asymmetric Temporal-Gradient QKV (ATG-QKV) mechanism strictly silences static redundancies by firing exclusively on motion gradients. To achieve true spatio-temporal sparsity, we introduce Lateral Spiking Topology Routing (LSTR) for zero-MAC conditional spike propagation strictly along the anatomical graph, coupled with a Spiking State-Space (S3) Engine that systematically sustains long-range causal memory to cure short-term amnesia. Extensive experiments on multiple large-scale datasets demonstrate that S3T-Former achieves highly competitive accuracy while theoretically consuming only a tiny fraction of the energy required by classic ANNs, establishing a definitive new state-of-the-art for energy-efficient neuromorphic action recognition.
\end{abstract}



\maketitle
\pagestyle{plain} 
\section{Introduction}
Skeleton-based action recognition is a fundamental pillar in multimedia understanding, pivotal for applications from embodied intelligent agents to immersive Extended Reality interactions \cite{liu2025systematic}. Raw skeleton data naturally provides rich multimodal features (e.g., spatial joints, structural bones) depicting human postures \cite{duan2022revisiting,ren2024survey}. Recently, Artificial Neural Networks (ANNs), particularly Graph Convolutional Networks (GCNs) \cite{ahmad2021graph,feng2022comparative} and Transformers \cite{plizzari2021skeleton,do2024skateformer,zhou2022hypergraph}, have achieved remarkable accuracy by fusing these multimodal features. However, extracting multi-perspective information heavily relies on dense, floating-point multiply-accumulate (MAC) operations. When processing continuous high-framerate sequences, this exorbitant energy consumption creates a severe bottleneck, prohibiting deployment on resource-constrained, battery-powered edge devices requiring "always-on" perception.

To address this energy crisis, Spiking Neural Networks (SNNs) have emerged as a promising neuromorphic paradigm, communicating via discrete binary spikes to replace power-hungry MACs with highly efficient accumulate (AC) operations\cite{ghosh2009spiking,khan2025spiking}. Recently, spiking-based Transformers have achieved remarkable success in various vision tasks\cite{zhou2022spikformer,yao2023spike,wang2023masked,lee2025spiking,yu2024spikingvit}. In the specific realm of skeleton-based action recognition, pioneering efforts have introduced Spiking Graph Convolutional Networks (Spiking-GCNs) to model human dynamics\cite{zheng2025mk,zheng2025signal}. However, to process complex multimodal skeletal features, existing methods often resort to computationally heavy fusion modules or non-sparse frequency domain transformations. These workarounds inadvertently compromise the extreme spatial-temporal sparsity and the pure additive (AC) computational advantage that are the core tenets of SNNs. Furthermore, in spatial topology modeling, they still inherit the dense matrix multiplication paradigm from traditional GCNs, failing to achieve conditional "on-demand" routing. Coupled with the intrinsic exponential decay ("short-term amnesia") of Leaky Integrate-and-Fire (LIF) neurons \cite{lansky2006parameters}, it remains a formidable challenge for SNNs to capture long-range temporal causality and perform true sparse reasoning.

Driven by the critical need for a fully sparse architecture, we propose the Spiking State-Space Topology Transformer (S3T-Former). Inspired by retinal ganglion cells that fire sparsely in response to spatio-temporal gradients, we introduce the Multi-Stream Anatomical Spiking Embedding (M-ASE) as a generalized kinematic differential operator. Because Leaky Integrate-and-Fire (LIF) neurons are intrinsically designed to spike on dynamic changes, M-ASE systematically computes multi-order dynamics (identity, spatial, and temporal gradients) to translate dense skeletal features into rich, highly sparse event streams. This differential extraction explicitly maximizes the representational capacity of a single base modality, establishing an exceptionally powerful single-stream architecture for complex action reasoning.

Within the core transformer blocks, we completely redesign the attention mechanism to maintain this extreme sparsity. First, mimicking the functional division of Magnocellular and Parvocellular cells in the primate visual system \cite{livingstone1988segregation, merigan1993parallel}, we introduce Asymmetric Temporal-Gradient QKV (ATG-QKV). By forcing queries and keys to fire exclusively on dynamic motion gradients while preserving static structural features for values, ATG-QKV strictly silences background redundancies and pushes event-driven sparsity to the extreme. Subsequently, to overcome the dense topology bottleneck, we propose Lateral Spiking Topology Routing (LSTR). Instead of computing power-hungry GCN aggregations or $O(V^2)$ dense attention matrices, LSTR elegantly transforms spatial graph modeling into zero-MAC conditional additions. By propagating bound Key-Value spikes purely on-demand along the anatomical graph, LSTR accurately captures structural synergies while maintaining strictly discrete, event-driven spatial sparsity.

Finally, to cure the intrinsic "short-term amnesia" of LIF neurons without resorting to non-sparse spectral workarounds, we seamlessly integrate the Spiking State-Space (S3) Engine. By constructing a linear-complexity temporal memory pool, the engine systematically accumulates topological spike features over extended time windows. Coupled with a continuous U-Readout mechanism that directly utilizes the final membrane potential to preserve sub-threshold decision confidence, S3T-Former is empowered with robust long-range memory capacity. Together, these innovations enable the network to execute complex, long-sequence spatio-temporal reasoning under an ultra-low power envelope.

The main contributions of this work are summarized as follows:
\begin{itemize}
    \item We propose \textbf{S3T-Former}, pioneering the first purely spike-driven Transformer architecture specifically designed for skeleton-based action recognition.
    
    \item We formulate \textbf{M-ASE} to translate multi-order biomechanical dynamics into rich event streams, maximizing the representational capacity of a single base modality without heavy intra-network fusion. Synergistically, we introduce \textbf{ATG-QKV} to force attention queries and keys to fire exclusively on motion gradients, strictly silencing static redundancies and pushing event-driven sparsity to the extreme.
    
    \item We introduce \textbf{LSTR} to elegantly transform $O(V^2)$ dense spatial graph modeling into zero-MAC conditional additions, and the \textbf{S3-Engine} to construct a linear-complexity temporal memory pool that cures SNN short-term amnesia, achieving robust long-range reasoning under strict spatio-temporal sparsity.
    
    \item Extensive experiments on three large-scale datasets demonstrate that S3T-Former establishes a new state-of-the-art for SNNs and successfully outperforms several established ANNs in pure accuracy, while consuming only a tiny fraction of their theoretical energy.
\end{itemize}
\section{Related Work}

\subsection{Skeleton-based Action Recognition}
Skeleton-based action recognition fundamentally relies on modeling joint kinematics. While early methods utilized RNNs\cite{liu2017skeleton,du2015hierarchical,liu2017skeleton-1} and CNNs\cite{zhang2019view,zhang2020semantics}, GCNs\cite{yan2018spatial,shi2019two,chen2021channel,xie2024dynamic} and Transformers\cite{zhang2022zoom,plizzari2021skeleton,do2024skateformer,zhou2022hypergraph} now dominate by effectively capturing spatio-temporal dependencies and global contexts. To maximize performance, fusing multiple modalities (joints, bones, motions) is standard practice\cite{chi2022infogcn,cheng2020skeleton,lee2023hierarchically, myung2024degcn}. However, these ANN paradigms rely heavily on power-hungry dense MAC operations, and complex multi-stream fusions further exacerbate computational burdens, prohibiting continuous efficient edge deployment.

\subsection{Spiking Neural Networks}
SNNs process information via discrete spikes, replacing power-hungry MACs with energy-efficient Accumulate (AC) operations \cite{ghosh2009spiking,khan2025spiking}. Recently, Spiking Transformers \cite{zhou2022spikformer,yao2023spike,wang2023masked,lee2025spiking,yu2024spikingvit} have demonstrated immense potential in vision tasks by combining global receptive fields with high sparsity. For skeleton data, pioneering works \cite{zheng2025mk} and \cite{zheng2025signal} explored Spiking GCNs. However, to handle complex dynamics, they often resort to heavy fusion modules, non-sparse spectral transforms, and dense matrix aggregations. These workarounds inadvertently compromise the extreme spatio-temporal sparsity inherent to neuromorphic computing.

\subsection{State-Space Models and Temporal Dynamics}
Modeling long-range dependencies is critical for continuous action reasoning. State-Space Models (SSMs)\cite{somvanshi2025s4,qu2024survey} efficiently capture ultra-long sequence contexts with linear complexity. In the neuromorphic domain, SNNs possess inherent temporal dynamics via the internal membrane potential of LIF neurons \cite{wu2018spatio, fang2021incorporating}. However, their intrinsic exponential leakage causes severe "short-term amnesia" over long skeleton sequences \cite{yao2022glif}. Sustaining long-term semantic causality without reverting to dense temporal convolutions or non-sparse spectral filters remains a critical open challenge.
\section{Method}
In this section, we present S3T-Former, a purely spike-driven architecture for energy-efficient skeleton action recognition. Beyond circumventing the exorbitant energy consumption of ANNs, S3T-Former is specifically formulated to resolve three critical bottlenecks in existing SNNs: weak perception of complex motion dynamics, pseudo-sparse graph topology, and short-term amnesia.
\begin{figure*}[t]
    \centering
    \includegraphics[width=0.95\textwidth]{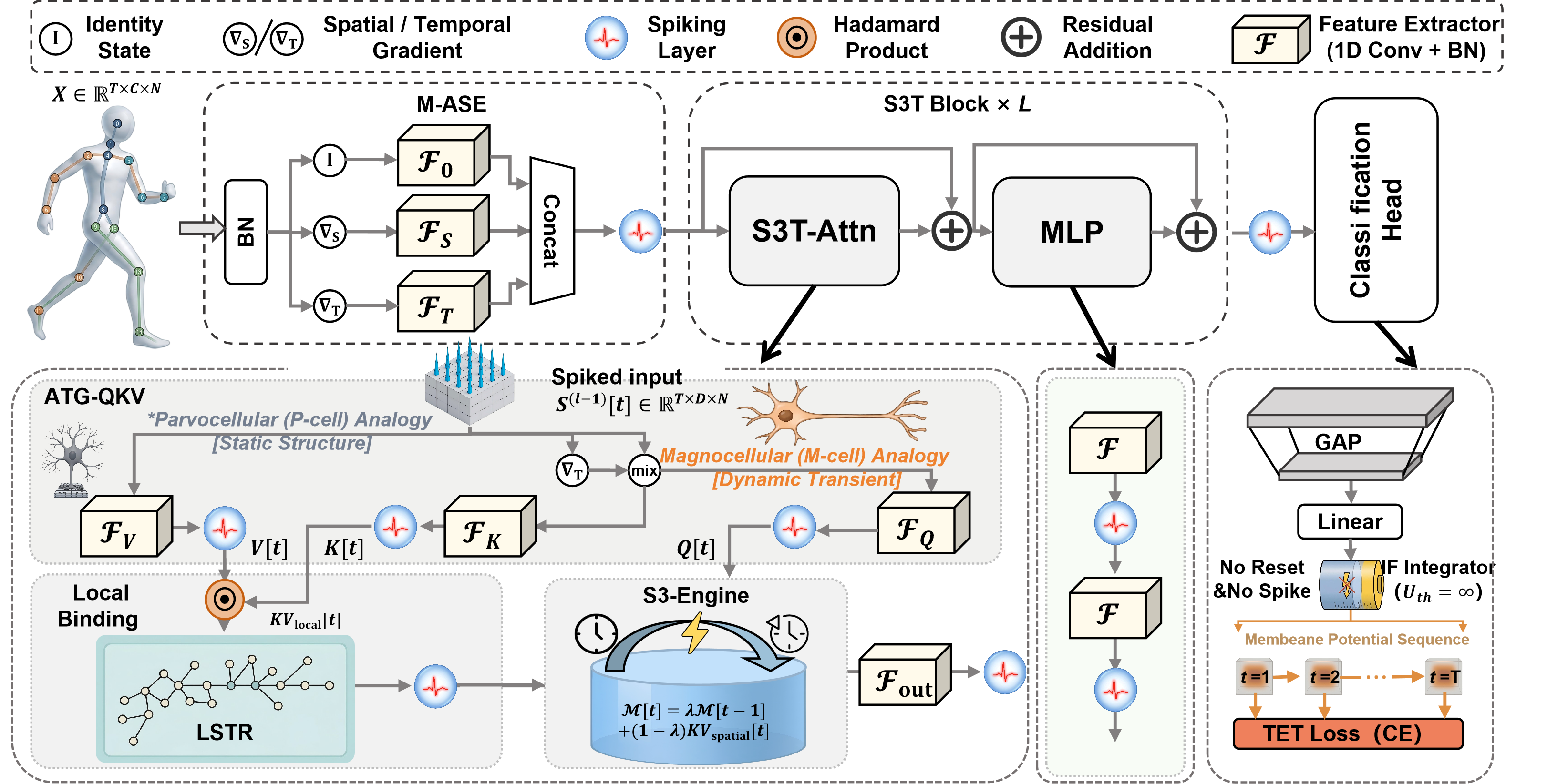} 
    \caption{Overall architecture of the proposed Spiking State-Space Topology Transformer (S3T-Former). The input skeletal coordinates are first processed by the Multi-Stream Anatomical Spiking Embedding (\textbf{M-ASE}) to generate rich, multi-order (identity, spatial, and temporal) event streams. The core S3T Block employs a Spiking State-Space Topology Attention (\textbf{S3T-Attn}) module and a \textbf{Spiking MLP}. A non-spiking \textbf{IF Integrator} ($U_{th}=\infty$) serves as a lossless membrane potential readout, optimized directly with Temporal Efficient Training (TET) loss.
    }
    \label{fig:overall_architecture}

\end{figure*}
\subsection{Preliminaries: Spiking Neuron Dynamics}
\label{sec:Pre}
Spiking Neural Networks process asynchronous event streams through the inherent temporal dynamics of spiking neurons. In this work, we adopt the widely-used Leaky Integrate-and-Fire (LIF) model as our foundational computing unit to ensure a purely event-driven architecture. The discrete-time dynamics of a LIF neuron at time step $t$ are formulated as follows:
\begin{equation}
    U[t] = \tau U[t-1] + I[t], \; S[t] = \Theta(U[t] - U_{th}), \; U[t] = U[t](1 - S[t]),
    \label{eq:lif_dynamics}
\end{equation}
where $I[t]$ represents the presynaptic input current, $U[t]$ denotes the internal membrane potential, and $\tau \in (0, 1)$ is the membrane leakage factor controlling the exponential decay of historical states. When $U[t]$ exceeds the predefined firing threshold $U_{th}$, the Heaviside step function $\Theta(\cdot)$ triggers the emission of a binary spike $S[t] \in \{0, 1\}$. Following the spike emission, a hard reset mechanism immediately returns the membrane potential to the resting state. By relying exclusively on these binary spike communications, LIF neurons replace MAC operations with highly efficient, AC operations throughout the network.
\subsection{Generalized Kinematic Differential Operator: M-ASE}
\label{sec:M-ASE}
Biological retinal ganglion cells encode visual information efficiently by firing sparsely in response to spatio-temporal gradients rather than absolute static intensities. Inspired by this neuromorphic mechanism, we propose the Multi-Stream Anatomical Spiking Embedding (M-ASE). Instead of functioning as a static linear projection, M-ASE is formulated as a generalized spatio-temporal kinematic differential operator. This design allows it to dynamically process any arbitrary base skeletal modality (e.g., joints, bones, or their temporal motions) and extract higher-order kinematic derivatives without heavy fusion architectures.

Given a generic base kinematic input $X \in \mathbb{R}^{T \times C_{in} \times N}$, where $T$, $C_{in}$, and $N$ denote the temporal length, coordinate channels, and number of nodes, M-ASE systematically evaluates its multi-order dynamics. We formalize this by computing the zero-order identity state $X^{(0)}$, the first-order temporal gradient $X^{(T)}$, and the first-order spatial gradient $X^{(S)}$:
\begin{equation}
    \begin{aligned}
        & X^{(0)}[t] = X[t], \quad X^{(T)}[t] = X[t] - X[t-1], \\
        & X^{(S)}[t, :, v_{tgt}] = X[t, :, v_{tgt}] - X[t, :, v_{src}],
    \end{aligned}
    \label{eq:m_ase_dynamics}
\end{equation}
where $(v_{src}, v_{tgt}) \in \mathcal{E}_{anat}$ defines the anatomically connected pairs directing from the human center to the extremities. Note that $X^{(T)}[0]$ is initialized to zero.

This generalized formulation is highly scalable and physically interpretable. For instance, when $X$ represents raw joint coordinates, $X^{(S)}$ and $X^{(T)}$ explicitly yield bone vectors and joint velocities. Crucially, when $X$ represents bone vectors, $X^{(S)}$ intrinsically captures structural curvatures (joint angles), and $X^{(T)}$ yields angular velocities.

Subsequently, these three heterogeneous streams are independently projected into respective sub-embedding spaces via 1D convolutions and batch normalization, denoted as $\mathcal{F}_0(\cdot)$, $\mathcal{F}_S(\cdot)$, and $\mathcal{F}_T(\cdot)$. Finally, the unified continuous representation is injected into a parametric LIF node to generate the initial highly sparse event stream $S_{in}[t] \in \{0, 1\}^{D \times N}$:
\begin{equation}
S_{in}[t] = \text{LIF} \left( \left[ \mathcal{F}_0(X^{(0)}[t]) \parallel \mathcal{F}_S(X^{(S)}[t]) \parallel \mathcal{F}_T(X^{(T)}[t]) \right] \right),
\label{eq:spike_embedding}
\end{equation}
where $D$ is the total embedding dimension. By explicitly calculating these higher-order differential states, M-ASE ensures that the subsequent S3T-Former perceives structural deformations and transient kinetics exactly as biological vision does—focusing entirely on active changes, thereby pushing the input spike sparsity to the extreme.

\subsection{Spiking State-Space Topology Block (S3T Block)}
\label{sec:S3T_Block}

To capture long-range spatio-temporal dependencies at minimal energy costs, we design the S3T Block as the core building block of the network. Taking the highly sparse event stream $S_{in}$ from the M-ASE module as its foundation, the S3T Block fundamentally decouples spatial aggregation and temporal integration. For a generic $l$-th block, let $S^{(l-1)} \in \{0, 1\}^{T \times D \times N}$ denote its input spike sequence (where $S^{(0)} = S_{in}$). We propose the \textbf{S3T} mechanism, which abandons traditional dense matrix multiplication. As illustrated in Fig. \ref{fig:overall_architecture}, the S3T module comprises three core innovations: ATG-QKV, LSTR, and the S3-Engine.

\subsubsection{Asymmetric Temporal-Gradient QKV (ATG-QKV)}
In the primate visual system, Magnocellular (M) cells are highly sensitive to dynamic motion transients, while Parvocellular (P) cells encode static structures \cite{livingstone1988segregation}. Existing spiking attention mechanisms disregard this biological division, generating Query and Key spikes uniformly across all nodes, which inherently causes severe computational redundancy. We introduce the ATG-QKV mechanism to force intention queries ($Q$ and $K$) to be dominated by physical motion gradients, while content representation ($V$) relies on static structural features.

Given the input spike sequence $S^{(l-1)}$ to the $l$-th layer, we extract its absolute temporal gradient $S_{grad}[t] = |S^{(l-1)}[t] - S^{(l-1)}[t-1]|$. To prevent completely static nodes from suffering ``feature starvation," we introduce a Soft-ATG strategy that blends the gradient with the original state. \textbf{For notational simplicity, we omit the layer index $(l)$ for all intermediate variables within the block}:
\begin{equation}
    S_{dyn}[t] = \alpha S_{grad}[t] + (1 - \alpha) S^{(l-1)}[t],
    \label{eq:soft_atg}
\end{equation}
where $\alpha \in \mathbb{R}^{D}$ is a learnable, channel-wise weighting parameter. Subsequently, these streams are independently projected into the attention space via distinct linear embedding functions, denoted as $\mathcal{F}_Q(\cdot)$, $\mathcal{F}_K(\cdot)$, and $\mathcal{F}_V(\cdot)$ (each comprising a 1D convolution followed by batch normalization). The query and key spikes are generated from the highly sparse dynamic stream $S_{dyn}$ through parametric LIF neurons, whereas the value spikes are derived from the original stable stream $S^{(l-1)}$:
\begin{equation}
\begin{aligned}
    Q[t] &= \text{LIF}_Q \big( \mathcal{F}_Q(S_{dyn}[t]) \big), \quad K[t] = \text{LIF}_K \big( \mathcal{F}_K(S_{dyn}[t]) \big) \\
    V[t] &= \text{LIF}_V \big( \mathcal{F}_V(S^{(l-1)}[t]) \big).
\end{aligned}
\label{eq:qkv_spike}
\end{equation}
For instance, during dynamic actions (e.g., running), rapidly moving limbs evoke intense motion gradients to trigger sparse $Q$ and $K$ spikes, while the relatively stationary torso provides stable $V$ spikes to anchor the spatial structure. This biologically plausible asymmetry intrinsically reduces the average spike firing rate by one to two orders of magnitude compared to standard symmetric architectures.

\subsubsection{Spiking State-Space Topology Attention}
After obtaining the $Q$, $K$, and $V$ spike streams, we abandon the $O(N^2 \times D)$ dense attention matrix computation. Instead, we perform an $O(N)$ spatio-temporal decoupled routing through three specialized steps:
\begin{figure}[t]
    \centering
    \includegraphics[width=\columnwidth]{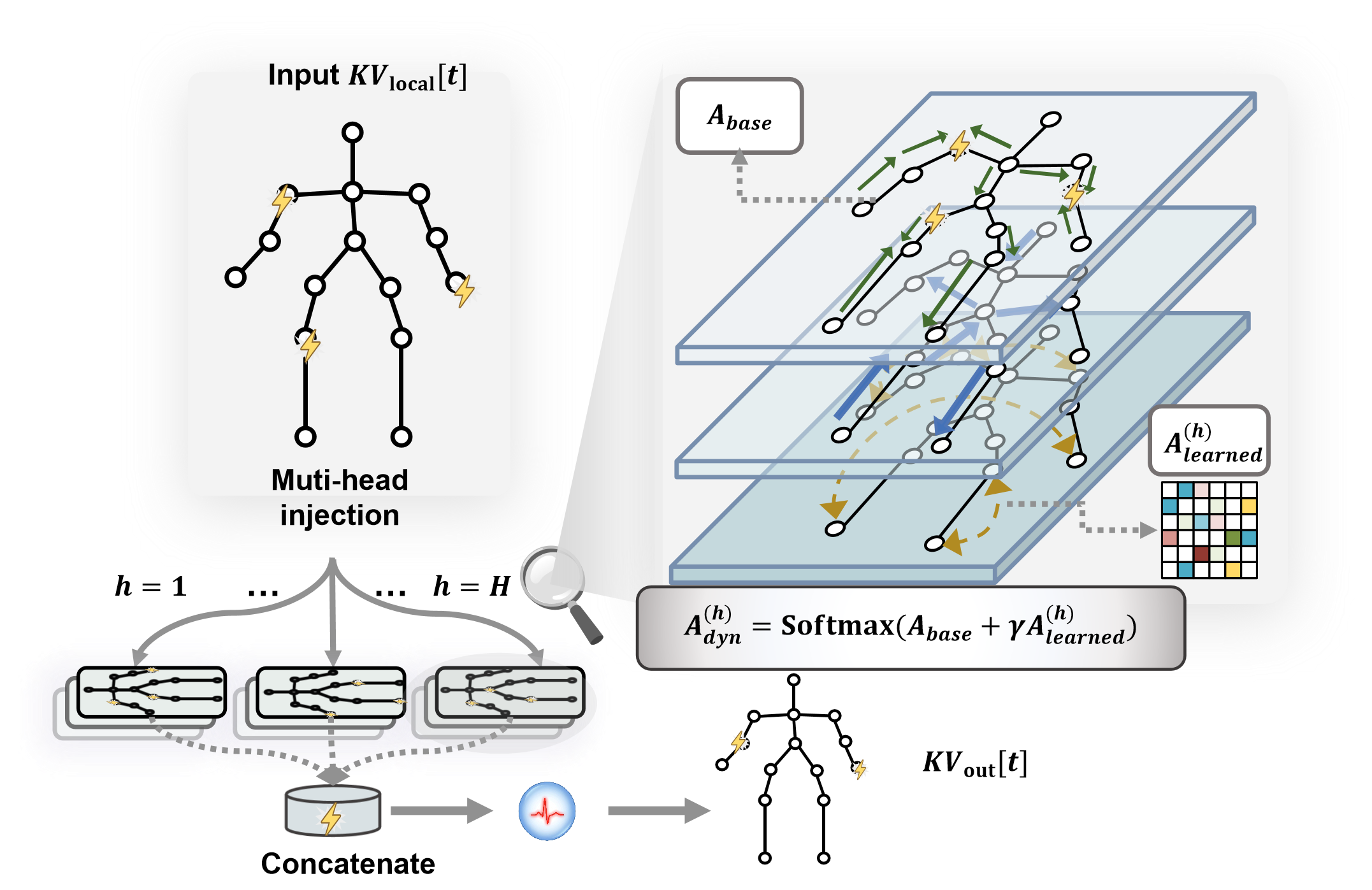}
    \caption{
        \textbf{Lateral Spiking Topology Routing (LSTR).} It decouples spatial anatomy into multi-head pathways, executing zero-MAC spatial feature broadcasts via conditional sparse additions.
    }
    \label{fig:lstr_micro}
\end{figure}

\textbf{Step A: Local Binding.} We first bind the dynamic intention with the static content within each node via a pure spike Hadamard product:
\begin{equation}
    KV_{\text{local}}[t] = K[t] \odot V[t]
    \label{eq:local_binding}
\end{equation}
Given the binary nature of spikes, this operation mathematically acts as a hard mask. Only nodes actively participating in motion are stamped and permitted to broadcast their features outward.

\textbf{Step B: Lateral Spiking Topology Routing (LSTR).} The masked $KV_{local}$ spikes are subsequently propagated to neighboring joints along the natural human anatomical graph, as illustrated in Fig. \ref{fig:lstr_micro}.
Formally, we model the human skeleton as a spatial graph $\mathcal{G} = (\mathcal{N}, \mathcal{E})$, where $\mathcal{N}$ represents the $N$ articulatory joints (nodes) and $\mathcal{E}$ denotes the physical bones (edges). The base topology matrix $A_{base} \in \mathbb{R}^{N \times N}$ is strictly derived from the normalized adjacency matrix of $\mathcal{G}$, explicitly encoding the hard physical connectivities (e.g., the hand is connected to the elbow, but not directly to the foot).

To capture diverse structural dependencies with data-driven flexibility, we decouple the spatial routing into $H$ independent attention heads. For each head $h \in \{1, \dots, H\}$, a head-specific dynamic topology matrix $\mathbf{A}_{dyn}^{(h)}$ is defined:
\begin{equation}
A_{dyn}^{(h)} = \text{Softmax}(A_{base} + \gamma A_{learned}^{(h)}),\label{eq:dynamic_topology}
\end{equation}
where $\mathbf{A}_{base}$ denotes the physical skeleton prior, $A_{learned}^{(h)}$ is a learnable parameter matrix initialized near zero to capture implicit, long-range synergistic dependencies, and $\gamma$ is scaling factor. The spatial routing is performed as:
\begin{equation}
KV_{\text{spatial}}^{(h)}[t] = A_{dyn}^{(h)} \cdot KV_{\text{local}}^{(h)}[t].
\label{eq:lstr_head}
\end{equation}
The routed features from all heads are concatenated and immediately re-spiked by a buffer LIF node to maintain the pure event-driven paradigm:
\begin{equation}
KV_{\text{out}}[t] = \text{LIF} \left( \left[ KV_{\text{spatial}}^{(1)}[t] \parallel \dots \parallel KV_{\text{spatial}}^{(H)}[t] \right] \right).\label{eq:lstr_integration}\end{equation}
Crucially, since $A_{base}$ and $A_{learned}^{(h)}$ are input-independent, the Softmax normalization is computed only once per inference rather than at every time step. Consequently, the weighted aggregation of binary spikes $KV_{\text{local}}^{(h)}[t]$ mathematically degenerates into hardware-friendly, conditional sparse ACs strictly along the topological edges, circumventing dense MAC operations entirely. This design allows LSTR to achieve high-precision spatial modeling with nearly zero additional computational overhead.

\begin{figure}[t]
    \centering
    \includegraphics[width=\columnwidth]{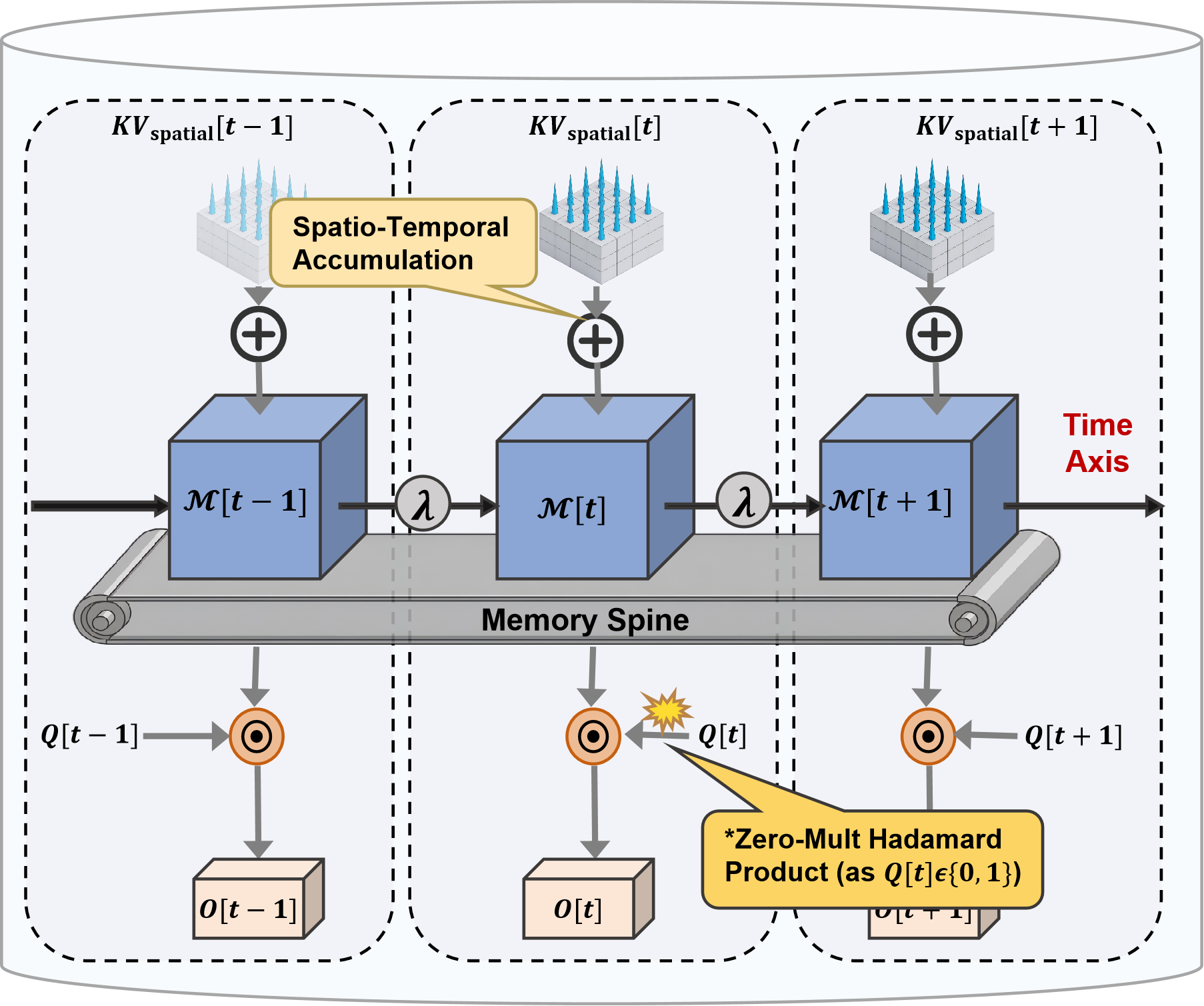}
    \caption{
        \textbf{Spiking State-Space (S3) Engine.} It constructs a linear-complexity temporal memory pool to integrate global spatio-temporal context without dense multiplications.
    }
    \label{fig:s3_engine_micro}
\end{figure}

\textbf{Step C: Spiking State-Space (S3) Engine.} To systematically cure the ``short-term amnesia" of LIF neurons without reverting to dense temporal convolutions, we construct a non-recurrent state-space memory pool along the temporal axis, as illustrated in Fig. \ref{fig:s3_engine_micro}. The learnable decay factor is bounded via
$\lambda = \text{Clamp}(\sigma(w_{decay}),$ $ 0.01, 0.99)$ to ensure stable long-term integration. The temporal memory state $\mathcal{M}[t]$ and the final attention output $O[t]$ are computed as:
\begin{align}
    \mathcal{M}[t] &= \lambda \odot \mathcal{M}[t-1] + (1 - \lambda) \odot KV_{\text{spatial}}[t], \label{eq:s3_memory}\\
    O[t] &= Q[t] \odot \mathcal{M}[t] \label{eq:s3_out}
\end{align}
This architecture enables the current query spike $Q[t]$ to directly retrieve the memory pool $\mathcal{M}[t]$, which inherently encapsulates the global historical topology. It bounds the overall sequence complexity to linear $O(T)$, offering a true spatio-temporal receptive field. The attention output is linearly projected, aggregated via an output LIF node, and passed through a standard Spiking MLP block via residual connections to accomplish channel mixing. To facilitate strict reproducibility and intuitively demonstrate the purely discrete, zero-MAC nature of our attention mechanisms, the complete PyTorch-like pseudocode for the S3T Block is provided in Appendix \ref{sec:app_pseudocode}.

\subsection{Overall Architecture and Membrane Readout}
\label{sec:Overall_Architecture}
Unlike vanilla Vision Transformers that rely heavily on explicit positional encodings, S3T-Former inherently dispenses with them. Spatially, the LSTR mechanism naturally embeds relative topology by routing spikes strictly along anatomical edges. Temporally, the causal formulation of the S3-Engine, combined with M-ASE temporal gradients, intrinsically captures the unidirectional flow of time. This implicit spatio-temporal encoding eliminates redundant learnable parameters while strictly adhering to event-driven neuromorphic principles.

After the final block, we perform Global Average Pooling (GAP) over the spatial dimension $N$ and aggregate the multi-person dimension to obtain a frame-level global representation $S^{(L)}_{\text{global}}[t] \in \mathbb{R}^{D}$. A fully connected layer then maps this representation to a continuous classification current $I_c[t] \in \mathbb{R}^{C_{classes}}$ at each time step.
To generate final predictions without the severe quantization errors of heuristic spike-counting, we propose \textbf{U-Readout}. It deploys a non-spiking Integrate-and-Fire (IF) node ($U_{th} = \infty$) as a lossless temporal accumulator. The final classification distribution is directly decoded from the terminal membrane potential $U_{out}[T]$. Furthermore, this continuous frame-by-frame integration inherently enables Temporal Efficient Training (TET) \cite{deng2022temporal}, applying dense objective supervision at every time step to facilitate optimal convergence and mitigate gradient vanishing in deep SNNs.
\section{Experiments}
We comprehensively evaluate S3T-Former on three widely adopted skeleton-based action recognition benchmarks: NTU RGB+D 60 \cite{shahroudy2016ntu}, NTU RGB+D 120 \cite{liu2019ntu}, and NW-UCLA \cite{wang2014cross}. Due to space constraints, detailed dataset protocols and implementation configurations are deferred to Appendices \ref{sec:app_datasets} and \ref{sec:app_implementation}. In this section, we first benchmark our model against state-of-the-art ANNs and SNNs to demonstrate its superior accuracy-efficiency trade-off. Subsequently, we provide extensive ablation studies and qualitative visualizations to dissect the individual contributions, routing interpretability, and extreme spiking sparsity of our core architectural designs.
\begin{table*}[t]
\caption{Comparison with prior ANN- and SNN-based models on NTU RGB+D, NTU RGB+D 120, and NW-UCLA datasets. `Mod.' denotes input modalities (J: Joint, B: Bone, JM: Joint Motion, BM: Bone Motion). The network type and architecture paradigm are denoted in brackets (e.g., [ANN-GCN], [SNN-Trans.]).}
\label{tab:sota_comparison}
\centering
\begin{tabularx}{\textwidth}{X c c c c c c c c c}
\toprule
Model & Mod. & $T$ & Param. & FLOPs 
& \multicolumn{2}{c}{NTU RGB+D} 
& \multicolumn{2}{c}{NTU RGB+D 120} 
& NW-UCLA\\

& & & (M)& (G) 
& Xs(\%) & Xv(\%) &Xs(\%) &Xt(\%) & (\%) \\
\midrule
Part-aware LSTM \cite{du2015hierarchical}         \hfill \textcolor{red!60}{\scriptsize [ANN-RNN]}   & J & - & -     & -     & 62.9 & 70.3 & 25.5 & 26.3 & - \\
ST-GCN \cite{yan2018spatial}                      \hfill \textcolor{red!60}{\scriptsize [ANN-GCN]}   & J & - & 3.10  & 3.48  & 81.5 & 88.3 & 70.7 & 73.2 & - \\
2S-AGCN \cite{shi2019two}                         \hfill \textcolor{red!60}{\scriptsize [ANN-GCN]}   & J+B & - & 3.48  & 37.32 & 88.5 & 95.1 & 82.5 & 84.3 & - \\
Shift-GCN \cite{cheng2020skeleton}                \hfill \textcolor{red!60}{\scriptsize [ANN-GCN]}   & All & - & - & 10.00 & 90.7 & 96.5 & 85.3 & 86.6 & 94.6 \\
MS-G3D \cite{liu2020disentangling}                \hfill \textcolor{red!60}{\scriptsize [ANN-GCN]}   & J+B & - & 3.19  & 48.88 & 91.5 & 96.2 & 86.9 & 88.4 & - \\
CTR-GCN \cite{chen2021channel}                    \hfill \textcolor{red!60}{\scriptsize [ANN-GCN]}   & All & - & 1.46  & 7.88  & 92.4 & 96.8 & 88.9 & 90.6 & 96.5 \\
\midrule
Hyperformer \cite{zhou2022hypergraph}            \hfill \textcolor{red!60}{\scriptsize [ANN-Trans.]}& J   & - & 2.60  & 14.80 & 90.7 & 95.1 & 86.6 & 88.0 & - \\
Hyperformer (4 ensemble) \cite{zhou2022hypergraph}\hfill \textcolor{red!60}{\scriptsize [ANN-Trans.]}& All & - & 2.60  & 59.20 & 92.9 & 96.5 & 89.9 & 91.3 & 96.7 \\
SkateFormer \cite{do2024skateformer}              \hfill \textcolor{red!60}{\scriptsize [ANN-Trans.]}& J   & - & 2.03  & 3.62  & 92.6 & 97.0 & 87.7 & 89.3 & - \\
SkateFormer (4 ensemble) \cite{do2024skateformer} \hfill \textcolor{red!60}{\scriptsize [ANN-Trans.]}& All & - & 2.03  & 14.48 & 93.5 & 97.8 & 89.8 & 91.4 & 98.3 \\
\midrule
Spikformer \cite{zhou2022spikformer}              \hfill \textcolor{blue!60}{\scriptsize [SNN-Trans.]}& J & 4 & 4.78  & 24.07 & 73.9 & 80.1 & 61.7 & 63.7 & 85.4 \\
Spike-driven Transformer \cite{yao2023spike}      \hfill \textcolor{blue!60}{\scriptsize [SNN-Trans.]}& J & 4 & 4.77  & 23.50 & 73.4 & 80.6 & 62.3 & 64.1 & 83.4 \\
Spike-driven Transformer V2 \cite{yao2024spike2}  \hfill \textcolor{blue!60}{\scriptsize [SNN-Trans.]}& J & 4 & 11.47 & 38.28 & 77.4 & 83.6 & 64.3 & 65.9 & 89.4 \\
Spiking Wavelet Transformer \cite{fang2025spiking}\hfill \textcolor{blue!60}{\scriptsize [SNN-Trans.]}& J & 4 & 3.24  & 19.30 & 74.7 & 81.2 & 63.5 & 64.7 & 86.7 \\
STAtten \cite{lee2025spiking}                     \hfill \textcolor{blue!60}{\scriptsize [SNN-Trans.]}& J & 4 & 3.19  & 18.78 & 72.8 & 79.7 & 60.3 & 61.7 & 82.8  \\
MK-SGN \cite{zheng2025mk}                         \hfill \textcolor{blue!60}{\scriptsize [SNN-GCN]}   & All & 4 & 2.17  & 7.84  & 78.5 & 85.6 & 67.8 & 69.5 & 92.3 \\
\midrule
Signal-SGN \cite{zheng2025signal}                 \hfill \textcolor{blue!60}{\scriptsize [SNN-GCN]}   & J   & 16 & 1.74  & 1.62  & 80.5 & 87.7 & 69.2 & 72.1 & 92.7 \\
Signal-SGN \cite{zheng2025signal}                 \hfill \textcolor{blue!60}{\scriptsize [SNN-GCN]}   & J+B & 16 & 1.74  & 3.24  & 82.5 & 89.2 & 71.3 & 74.2 & 93.1 \\
Signal-SGN (4 ensemble) \cite{zheng2025signal}    \hfill \textcolor{blue!60}{\scriptsize [SNN-GCN]}   & All & 16 & 1.74  & 6.48  & 86.1 & 93.1 & 75.3 & 77.9 & 95.9 \\
\midrule
\rowcolor[gray]{0.95}
S3T-Former (D=256)                                \hfill \textbf{\textcolor{blue!80}{\scriptsize [SNN-Trans.]}}  & J   & 8  & 4.80  & 0.957  & 82.83  & 89.65 & 74.01  & 78.98 & 93.1 \\
\rowcolor[gray]{0.95}
S3T-Former (D=256)                                \hfill \textbf{\textcolor{blue!80}{\scriptsize [SNN-Trans.]}}  & J   & 16 & 4.80  & 1.914  & 83.61  & 90.42 & 74.78  & 79.45 & 93.8 \\
\rowcolor[gray]{0.95}
S3T-Former (D=256)                                \hfill \textbf{\textcolor{blue!80}{\scriptsize [SNN-Trans.]}}  & J   & 32 & 4.80  & 3.829  & 84.50  & 91.15 & 75.58  & 80.01 & 94.3 \\
\rowcolor[gray]{0.95}
S3T-Former (D=384)                                \hfill \textbf{\textcolor{blue!80}{\scriptsize [SNN-Trans.]}}  & J   & 16 & 10.73 & 4.287  & 84.94  & 91.20 & 75.76  & 80.24 & 94.8 \\
\rowcolor[gray]{0.95}
S3T-Former (D=384)                                \hfill \textbf{\textcolor{blue!80}{\scriptsize [SNN-Trans.]}}  & J   & 32 & 10.73 & 8.574  & 85.12  & 91.5 & 76.12  & 80.65 & 95.1 \\
\rowcolor[gray]{0.95}
S3T-Former (D=384)                                \hfill \textbf{\textcolor{blue!80}{\scriptsize [SNN-Trans.]}}  & J+B & 16 & 10.73 & 8.574  & 86.30  & 93.15 & 79.36  & 82.15 & 95.8 \\
\rowcolor[gray]{0.9}
\textbf{S3T-Former (D=384, Ensemble)}             \hfill \textbf{\textcolor{blue!80}{\scriptsize [SNN-Trans.]}}  & \textbf{All} & \textbf{16} & \textbf{10.73} & \textbf{17.14} & \textbf{87.35} & \textbf{94.05} & \textbf{81.68} & \textbf{84.12} & \textbf{96.6} \\
\bottomrule
\end{tabularx}
\vspace{1ex}
\footnotesize Param. represents the model size. FLOPs represent theoretical floating-point operations. For SNN methods, the $T$ column explicitly denotes the neuromorphic simulation time window. We omit $T$ for ANN models to avoid conceptual confusion with their raw input frame length. "All" refers to the fusion of J, B, JM, and BM modalities. \textbf{Evaluation protocols are denoted as Xs (Cross-Subject), Xv (Cross-View), and Xt (Cross-Setup).}
\end{table*}
\subsection{Comparison with State-of-the-Art Architectures} 

We benchmark S3T-Former against state-of-the-art architectures, spanning advanced Artificial Neural Networks (ANNs) and Spiking Neural Networks (SNNs). 

Crucially, compared to the current neuromorphic state-of-the-art, Signal-SGN \cite{zheng2025signal}, our S3T-Former demonstrates absolute superiority. Under the exact same simulation window ($T=16$) and single joint ($J$) modality, our base model ($D=256$) achieves 83.61\% on NTU RGB+D 60 (X-Sub), outperforming Signal-SGN (80.5\%) by a significant +3.11\% margin. Scaling the embedding dimension ($D=384$) further widens this gap to +4.44\% (84.94\%). This dominance is even more pronounced on the highly challenging NTU RGB+D 120 dataset, where our ensemble model achieves 81.68\% (X-Sub), completely eclipsing the Signal-SGN ensemble (75.3\%) by an astounding +6.38\%. This compelling evidence proves that our purely spike-driven topology attention and state-space memory systematically capture complex, long-range temporal dynamics far better than existing Spiking GCNs. Furthermore, S3T-Former shatters the performance ceiling of early Spiking Transformers (e.g., Spikformer\cite{zhou2022spikformer}, Spike-driven Transformer V2\cite{yao2024spike2}), which plateaued around 73\%-77\% on NTU-60.

When compared against traditional ANNs, S3T-Former showcases an unparalleled accuracy-efficiency trade-off. While recent ANN models (e.g., Hyperformer\cite{zhou2022hypergraph} , SkateFormer\cite{do2024skateformer}) achieve impressive peak accuracy, they inherently suffer from exorbitant computational burdens driven by power-hungry floating-point MAC operations. For instance, SkateFormer's ensemble requires 14.48G FLOPs, and Hyperformer demands up to 59.20G. In stark contrast, our single-stream S3T-Former ($D=256$) requires merely 1.91G theoretical FLOPs.  More importantly, a direct FLOP-to-FLOP comparison severely underestimates our neuromorphic advantage. As detailed in Appendix \ref{sec:app_energy_sparsity}, S3T-Former executes the vast majority of its operations as binary, sparse Synaptic Operations (SOPs). By replacing dense MACs with AC operations and perfectly silencing static spatial nodes via the ATG-QKV mechanism, conservative estimations guarantee that our model consumes less than 10\% of the physical energy required by standard ANN counterparts. Despite this extreme power reduction, S3T-Former achieves the rare neuromorphic feat of surpassing several established ANN models in pure accuracy (e.g., our 84.94\% explicitly beats ST-GCN's 81.5\% on NTU-60 X-Sub), establishing a definitive new state-of-the-art for energy-efficient action recognition \cite{yan2018spatial}.
\begin{table}[t]
\caption{Progressive core build-up ablation study of S3T-Former. UR: U-Readout, S3: S3-Engine, LS: LSTR, MA: M-ASE, AQ: ATG-QKV. FLOPs are reported for a single person.}
\label{tab:core_ablation}
\centering
\setlength{\tabcolsep}{3.5pt} 
\resizebox{\columnwidth}{!}{%
\begin{tabular}{c c c c c c c c c} 
\toprule
\multirow{2}{*}{Step} & \multicolumn{5}{c}{Components} & Param. & FLOPs & Acc \\
\cmidrule(lr){2-6}
& UR & S3 & LS & MA & AQ & (M) & (G) & (\%) \\
\midrule
1 (Baseline)& & & & & & 4.77 & 1.91 & 76.32 \\
2 & \checkmark & & & & & 4.77 & 1.91 & 77.15 \\
3 & \checkmark & \checkmark & & & & 4.77 & 1.91 & 78.68 \\
4 & \checkmark & \checkmark & \checkmark & & & 4.78 & 1.91 & 80.24 \\
5 & \checkmark & \checkmark & \checkmark & \checkmark & & 4.78 & 1.91 & 82.77 \\
\rowcolor[gray]{0.95}
\textbf{6} & \checkmark & \checkmark & \checkmark & \checkmark & \checkmark & \textbf{4.78} & \textbf{1.91} & \textbf{83.80} \\
\bottomrule
\end{tabular}%
}
\vspace{-2ex} 
\end{table}
\subsection{Ablation Studies}
To deeply analyze the underlying mechanisms of S3T-Former and validate the efficacy of our architectural designs, we conduct extensive ablation studies on the NTU RGB+D 60 (Cross-Subject) dataset. Our evaluation spans from macroscopic to microscopic dimensions: first, we progressively integrate our core modules to validate their synergistic effects and overall performance gains (Sec.\ref{sec:4.2.1}); second, we explore the lateral topology routing of the LSTR module to verify the necessity of fusing physical priors with data-driven routing (Sec.\ref{sec:4.2.2}); Finally, we dissect the temporal state-space memory and continuous readout strategies of the S3-Engine (Sec.\ref{sec:4.2.3}).
\subsubsection{Progressive Core Build-up}
\label{sec:4.2.1}
We evaluate the individual and synergistic contributions of our proposed modules by progressively integrating them into a "Vanilla Spiking Transformer" baseline. This baseline is constructed following the standard Spikformer architecture, utilizing LIF neurons and dense attention on a single joint stream. Notably, to ensure a rigorous "apples-to-apples" ablation, we equip this baseline with our optimized training recipe (e.g., an extended simulation window of $T=16$ and Temporal Efficient Training). Consequently, it establishes a solid foundation of 76.32\% Top-1 accuracy—significantly outperforming the skeleton-adapted Spikformer baseline ($73.9\%$, $T=4$) reproduced in prior works \cite{zheng2025mk}.
As shown in Table \ref{tab:core_ablation}, building upon this strong baseline, the integration of U-Readout (Step 2) and the S3-Engine (Step 3) enhances performance to 78.68\% (+2.36\%), demonstrating the critical role of continuous membrane potential dynamics and temporal state-space trajectories in mitigating the "short-term amnesia" inherent in traditional SNNs. The addition of LSTR (Step 4)further boosts accuracy to 80.24\% by enforcing anatomically-guided sparse routing. A significant leap to 82.77\% occurs with M-ASE (Step 5), proving that multi-order kinematic representations—joints, bones, and motion—fully unlock the potential of topological routing. Finally, our full S3T-Former (Step 6) achieves a peak accuracy of 83.80\% via the ATG-QKV mechanism, which optimizes attention logic through asymmetric temporal gradients. Remarkably, this substantial +7.48\% overall improvement is achieved with a practically imperceptible computational footprint (adding merely 0.01M parameters with mathematically negligible FLOPs overhead), showcasing a successful decoupling of accuracy enhancement from computational inflation.
\begin{table}[h]
\caption{Comprehensive ablation of LSTR routing strategies, multi-head configurations ($H$), and scaling factors ($\gamma$).}
\label{tab:lstr_study}
\centering
\setlength{\tabcolsep}{8pt}
\begin{tabular}{l c c c}
\toprule
Routing Strategy & Heads & Factor& Acc \\
& $H$ &$\gamma$ &  (\%) \\
\midrule
Dense (No Prior) & - & - & 78.68 \\
Physical Only (Static) & - & - & 79.12 \\
\midrule
LSTR (Hybrid) & Shared ($H=1$) & 0.5 & 81.45 \\
\midrule
LSTR (Hybrid) & Split ($H=4$) & 0.1 & 81.75 \\
LSTR (Hybrid) & Split ($H=4$) & 1.0 & 81.20 \\
LSTR (Hybrid) & Split ($H=4$) & 0.5 & 82.91 \\
\midrule
LSTR (Hybrid) & Split ($H=8$) & 0.1 & 82.34 \\
LSTR (Hybrid) & Split ($H=8$) & 1.0 & 81.92 \\
\rowcolor[gray]{0.95}
\textbf{LSTR (Hybrid)} & \textbf{Split ($H=8$)} & \textbf{0.5} & \textbf{83.80} \\
\bottomrule
\end{tabular}
\end{table}
\subsubsection {Exploration of LSTR}
\label{sec:4.2.2}
We evaluate the LSTR module against Dense (all-to-all) and Static Physical baselines (Table \ref{tab:lstr_study}), demonstrating that our hybrid approach effectively balances structural constraints with global context to filter redundant spatial noise. We further ablate the multi-head configurations ($H$) and the scaling factor ($\gamma$). Progressively increasing $H$ to 8 yields consistent gains, proving that independent heads effectively specialize in distinct kinematic sub-structures to enhance representational diversity. Simultaneously, a moderate scaling factor of $\gamma=0.5$ emerges as the optimal sweet spot: a smaller value (0.1) overly restricts routing to rigid skeletal links, while a larger value (1.0) diminishes the stabilizing physical prior and amplifies spatial noise. Ultimately, configuring LSTR with $H=8$ and $\gamma=0.5$ perfectly grounds the network in physical reality while maintaining vital data-driven adaptability.
\subsubsection{Synergy of S3-Engine and U-Readout}
\label{sec:4.2.3}

\begin{table}[h]
\caption{Ablation of S3-Engine and output readout strategies.}
\label{tab:s3_study}
\centering
\setlength{\tabcolsep}{6pt}
\begin{tabular}{l c c c}
\toprule
Temporal Memory & Decay Factor & Readout Type & Acc \\
 & $\lambda$ & &  (\%) \\
\midrule
None & - & Spike-rate & 76.32 \\
None & - & U-Readout & 77.15 \\
\midrule
S3-Engine & Learnable & Spike-rate & 78.68 \\
S3-Engine & Fixed (0.5) & U-Readout & 79.42 \\
S3-Engine & Linear & U-Readout & 81.18 \\
\rowcolor[gray]{0.95}
\textbf{S3-Engine} & \textbf{Learnable} & \textbf{U-Readout} & \textbf{83.80} \\
\bottomrule
\end{tabular}
\end{table}
We investigate the synergy between the supplemental S3-Engine and the U-Readout decoding strategy. As shown in Table \ref{tab:s3_study}, the baseline without the S3-Engine yields 76.32\% under standard Spike-rate decoding, and upgrading solely to U-Readout provides only a marginal gain (77.15\%). Furthermore, even when introducing a fully learnable S3-Engine to preserve historical trajectories, performance remains bottlenecked at 78.68\% if forced through a discrete Spike-rate classifier, as the rich continuous temporal context is severely degraded by non-differentiable quantization errors. The network's true potential is unlocked only when these mechanisms are synergistically combined: by employing U-Readout to directly utilize the final accumulated membrane potential $U_{out}[T]$, the model completely bypasses quantization loss to fully exploit the sophisticated temporal states, achieving a peak accuracy of 83.80\%. This demonstrates a strict co-dependency where the S3-Engine safeguards long-range temporal causality, and U-Readout ensures absolute information fidelity during final decision decoding.

\subsection{Qualitative Results}
\subsubsection{Visualization of Topological Routing} 
\begin{figure}[t]
    \centering
    \begin{subfigure}[b]{0.15\textwidth}
        \centering
        \includegraphics[width=\textwidth]{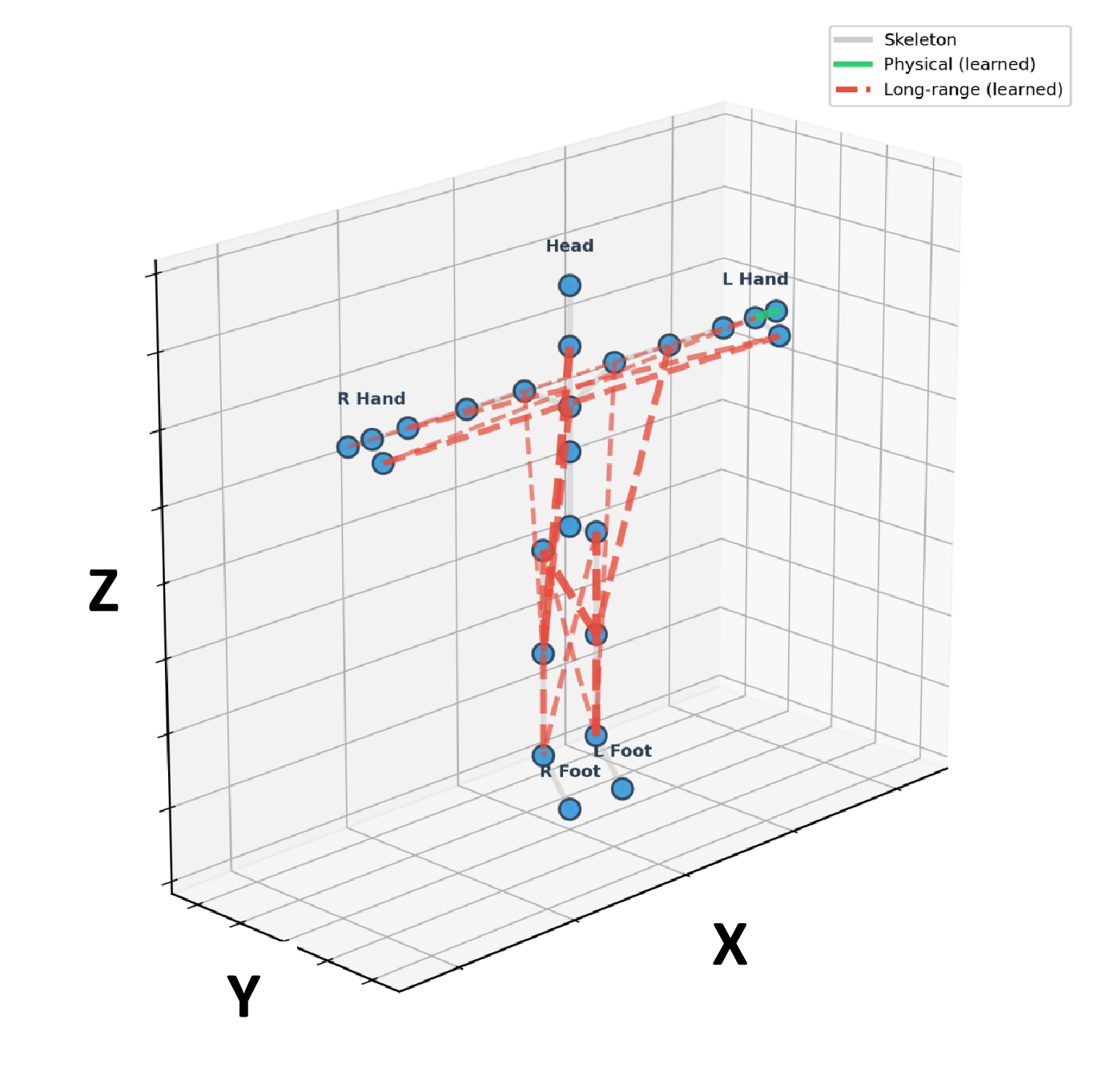} 
        \caption{$L=1, h=8$}
        \label{fig:topo_L1}
    \end{subfigure}
    \hfill
    \begin{subfigure}[b]{0.15\textwidth}
        \centering
        \includegraphics[width=\textwidth]{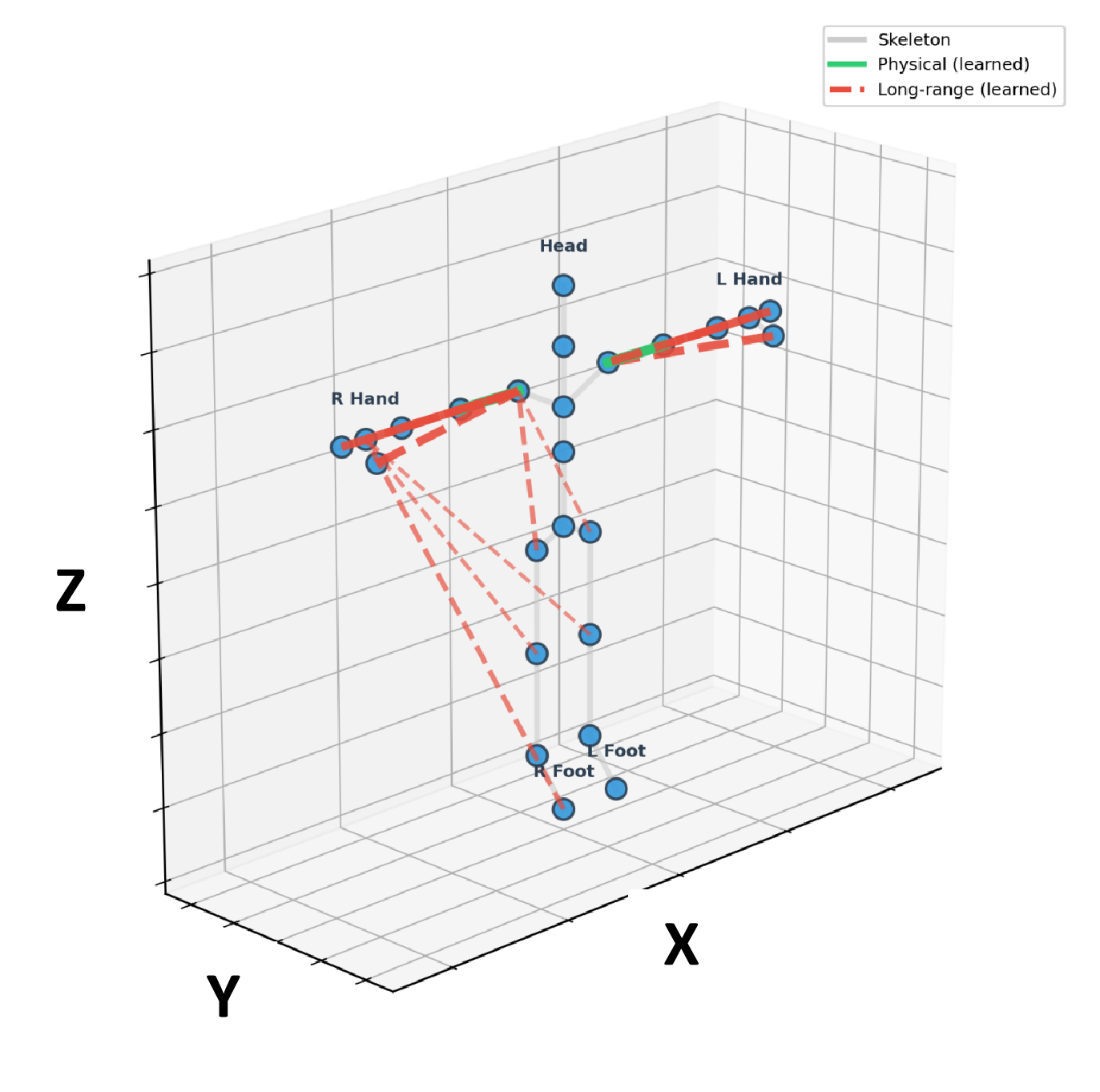} 
        \caption{$L=3, h=8$}
        \label{fig:topo_L3}
    \end{subfigure}
    \hfill
    \begin{subfigure}[b]{0.15\textwidth}
        \centering
        \includegraphics[width=\textwidth]{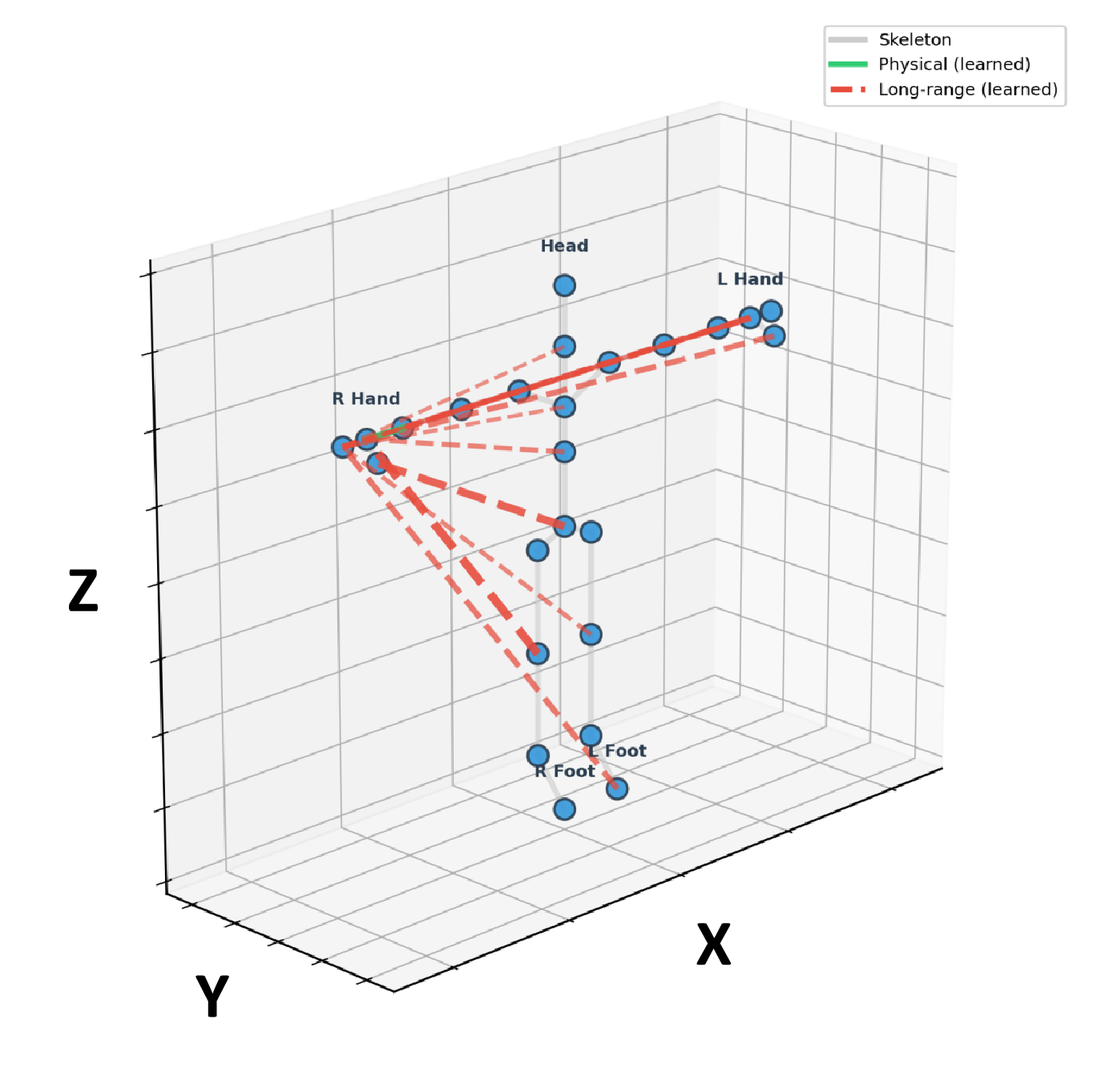} 
        \caption{$L=6, h=8$}
        \label{fig:topo_L6}
    \end{subfigure}
    \caption{Visualization of the dynamic topology matrix $\mathbf{A}_{dyn}^{(h)}$ (Eq. \ref{eq:dynamic_topology}) across different network depths for head $h=8$.}
    \label{fig:topo_evolution}
\end{figure}

To illustrate how independent heads specialize in distinct sub-structures, we visualize the dynamic topology $\mathbf{A}_{dyn}^{(8)}$ for a representative head ($h=8$, Fig. \ref{fig:topo_evolution}). It reveals a profound layer-wise convergence: shallow layers ($L=1$) aggressively explore global contexts, expanding receptive fields but introducing redundancy. As representations propagate ($L=3$), the module actively prunes irrelevant peripheral links. Ultimately, deep layers ($L=6$) sharpen the topology into a focused semantic cluster (e.g., coordinating the dominant arm and torso), demonstrating LSTR's ability to filter spatial noise and distill essential kinematic signatures.
\begin{figure}[htbp]
  \centering
  \begin{subfigure}{0.48\linewidth}
    \centering
    \includegraphics[width=\linewidth]{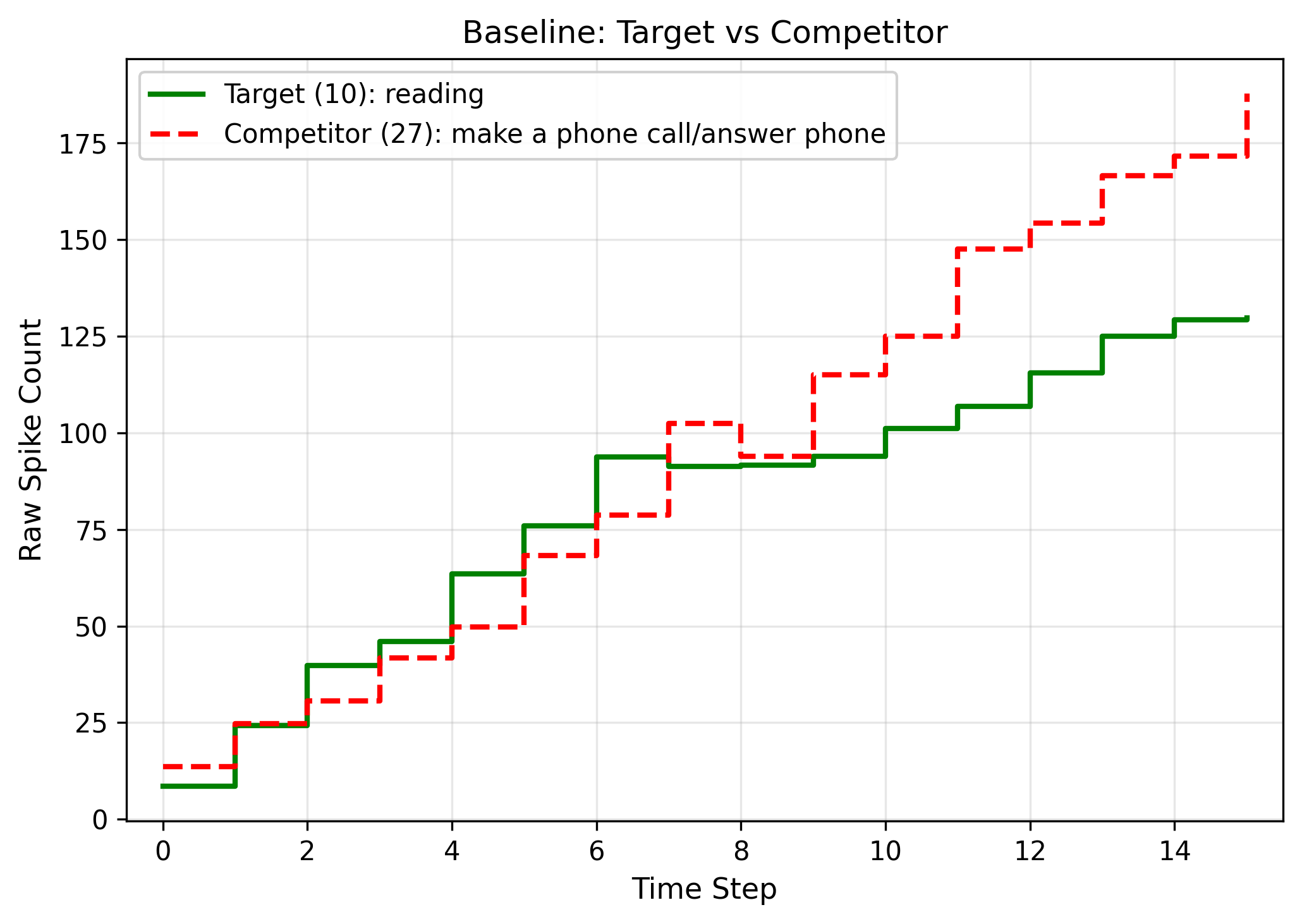}
    \caption{\small Baseline}
    \label{fig:target_baseline}
  \end{subfigure}
  \hfill 
  \begin{subfigure}{0.48\linewidth}
    \centering
    \includegraphics[width=\linewidth]{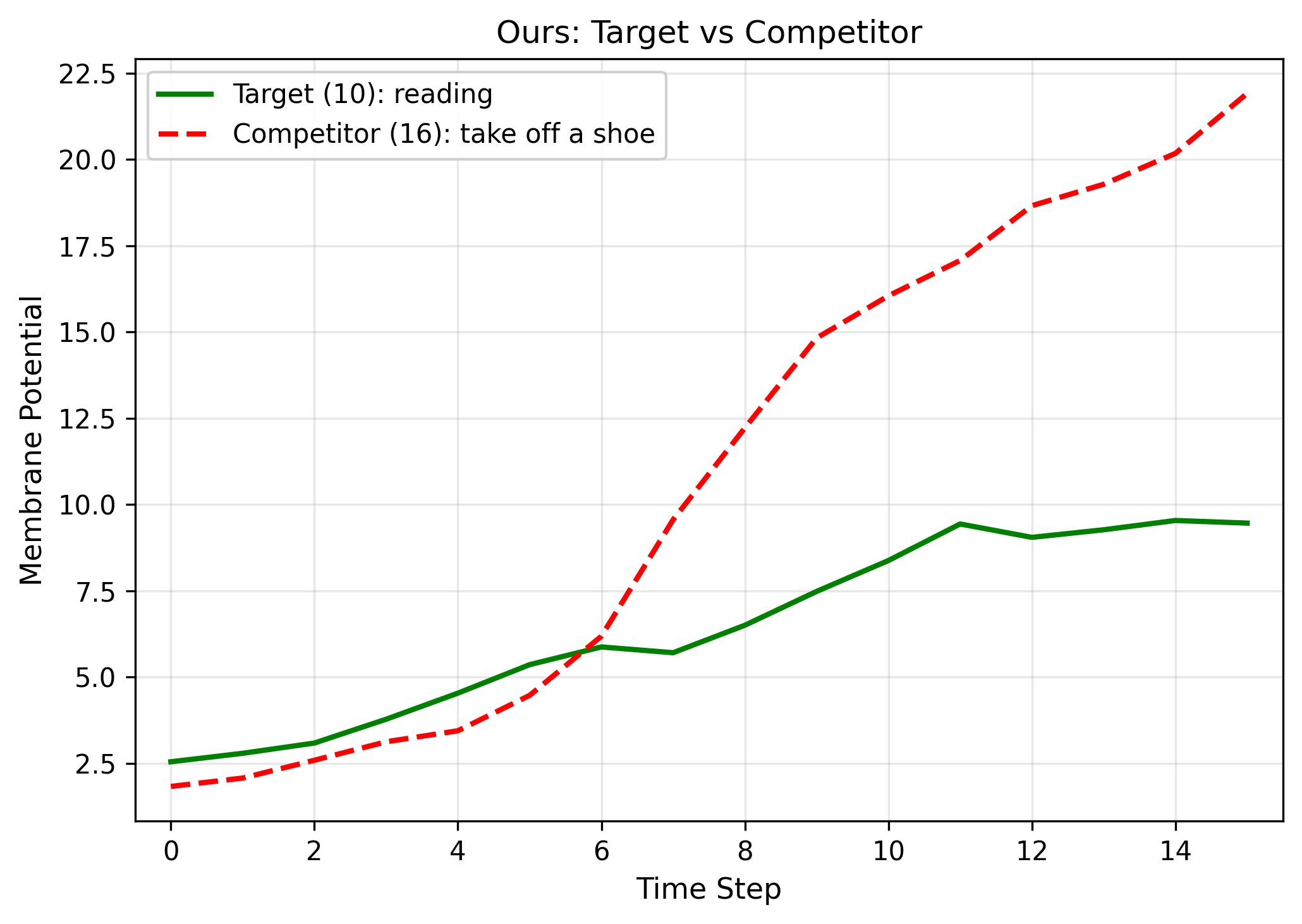}
    \caption{\small Ours}
    \label{fig:target_ours}
  \end{subfigure}
  
  \caption{Target vs. Competitor dynamics. U-Readout avoids the staircase effect of Spike Counting.}
  \label{fig:target_vs_competitor}
\end{figure}

\begin{figure}[htbp]
  \centering
  \begin{subfigure}{0.48\linewidth}
    \centering
    \includegraphics[width=\linewidth]{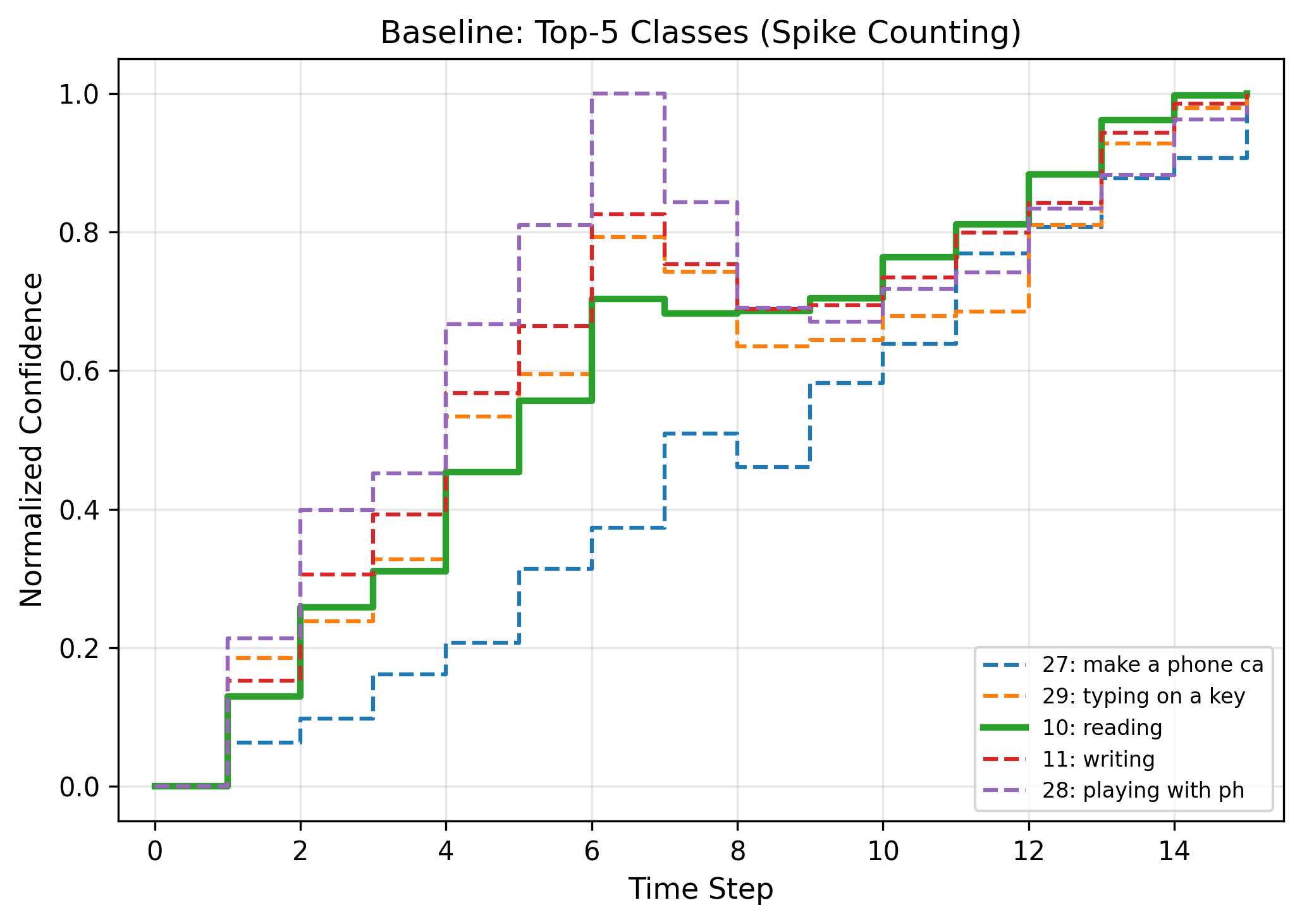}
    \caption{\small Baseline}
    \label{fig:top5_baseline}
  \end{subfigure}
  \hfill
  \begin{subfigure}{0.48\linewidth}
    \centering
    \includegraphics[width=\linewidth]{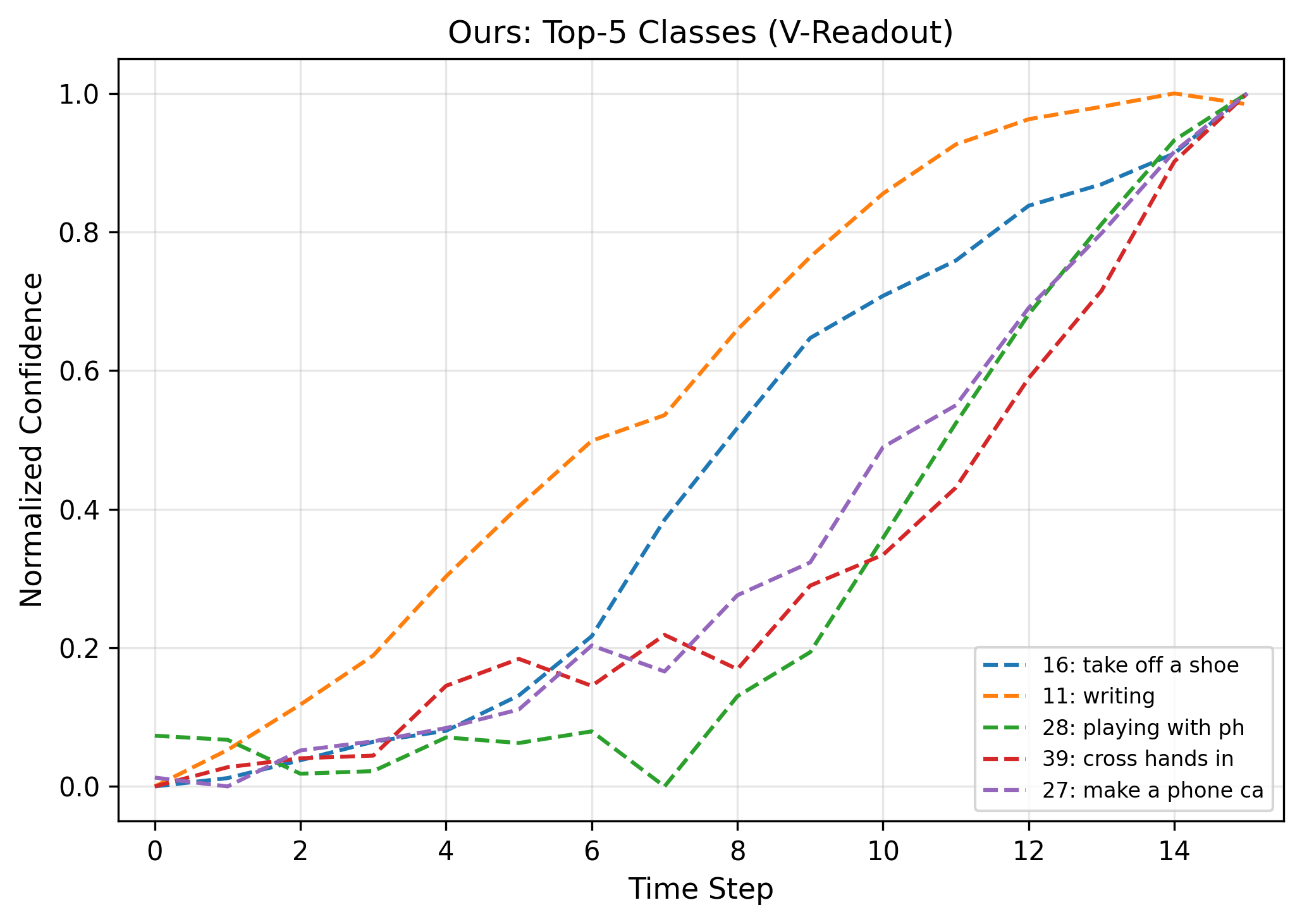}
    \caption{\small Ours}
    \label{fig:top5_ours}
  \end{subfigure}
  
  \caption{Top-5 class progression. U-Readout provides smoother, more distinguishable trajectories.}
  \label{fig:top5_classes}
\end{figure}
\subsubsection{Output Dynamics Analysis.} As shown in Fig. \ref{fig:target_vs_competitor} and \ref{fig:top5_classes}, the baseline spike-counting exhibits a discrete "staircase effect," suffering from severe sub-threshold quantization loss. Conversely, U-Readout generates smooth confidence trajectories from membrane potentials, preserving fine-grained temporal evolution. This establishes a distinctly clearer margin between the target class and its strongest competitor, overcoming the resolution limitations of traditional SNN decoding.

\subsubsection{Spatio-Temporal Sparsity Analysis}
\begin{figure}[t]
    \centering
    \includegraphics[width=0.5\textwidth]{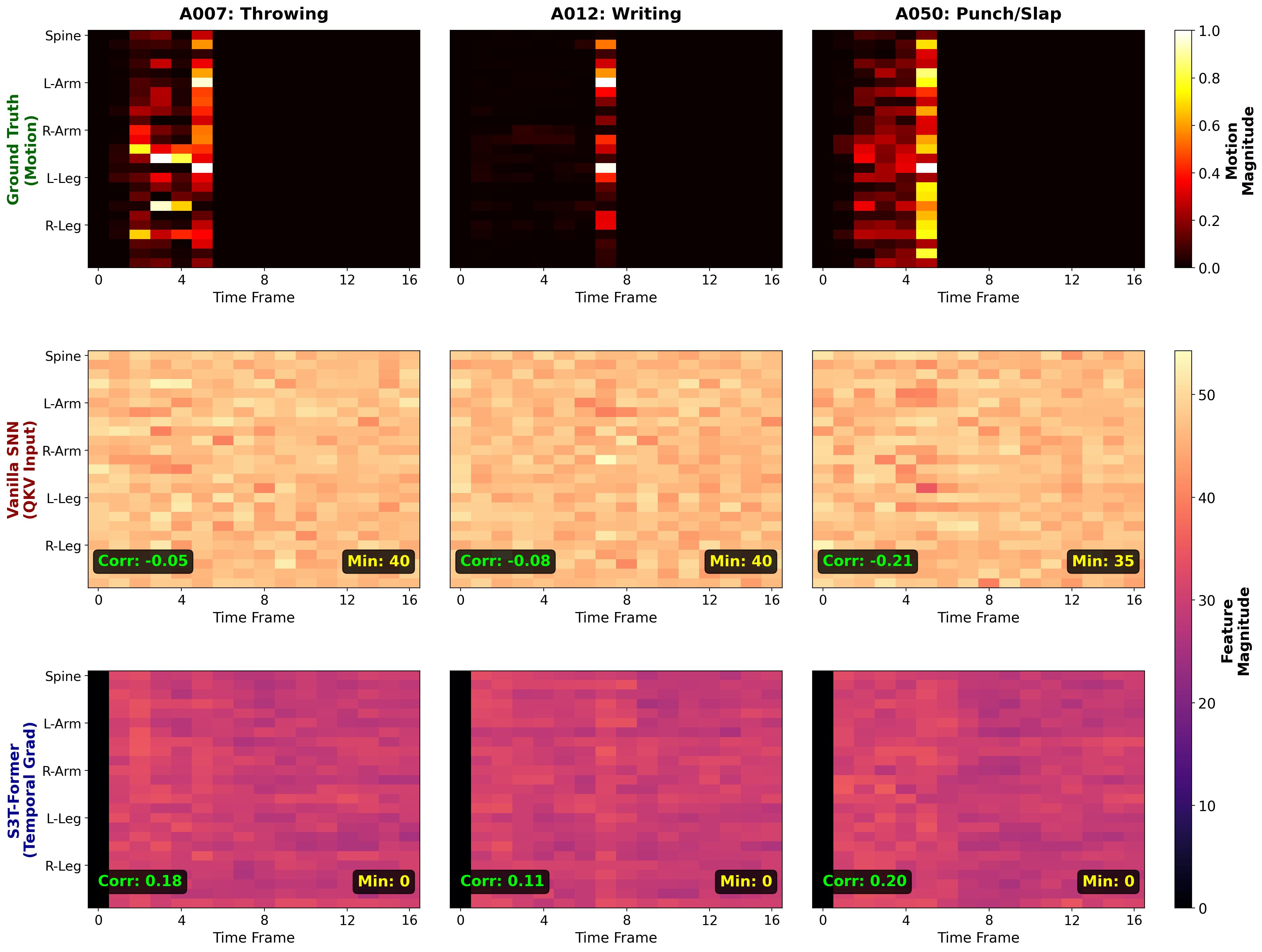}
    \caption{Visualization of spatio-temporal activation maps across three distinct action classes.}
    \label{fig:sparsity_analysis}
\end{figure}
Unlike vanilla SNNs that maintain energy-wasting background noise for static joints (yielding negative motion correlations, e.g., -0.21), S3T-Former completely silences static frames (Fig. \ref{fig:sparsity_analysis}). Activations peak exclusively during meaningful joint displacements, yielding consistent positive correlations across diverse actions. This confirms ATG-QKV successfully eradicates spatial redundancy, dynamically allocating computations strictly to salient kinetic events.
\begin{figure}[t]
    \centering
    \includegraphics[width=0.5\textwidth]{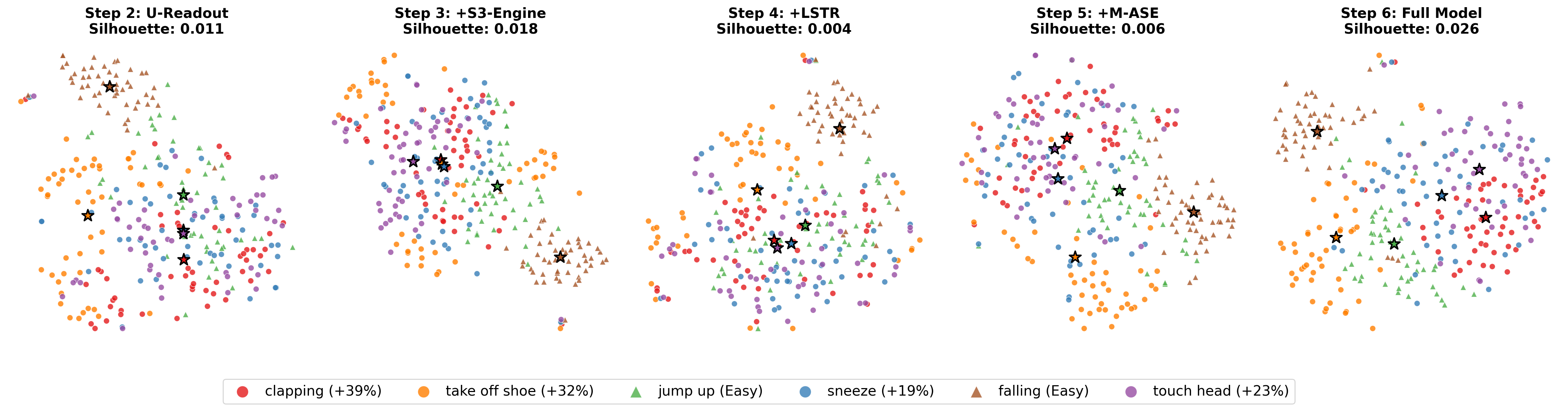} 
    \caption{t-SNE visualization of learned feature distributions for visually confusing action classes.}
    \label{fig:tsne_ablation}
\end{figure}
\subsubsection{Feature Representation Discriminability.}
We visualize feature distributions using t-SNE (Fig. \ref{fig:tsne_ablation}) for four visually confusing classes with similar localized movements. S3T-Former successfully maps these challenging actions into well-separated clusters with tight intra-class variance. Remarkably, even highly ambiguous actions like "Reading" and "Writing"—sharing almost identical minimal facial-proximal dynamics—are clearly disentangled, confirming robust semantic decision boundaries. Due to space constraints, further qualitative analyses—including the full confusion matrix, layer-wise semantic sparsity funnels, and a detailed per-class performance breakthrough analysis—are deferred to Appendices \ref{sec:app_visuals}.
\section{Conclusion}
We propose S3T-Former, the first purely spike-driven Transformer for energy-efficient skeleton action recognition. To circumvent heavy multimodal fusion, our Multi-Stream Anatomical Spiking Embedding (M-ASE) translates multi-order kinematics into rich event streams. Synergistically, the Asymmetric Temporal-Gradient QKV (ATG-QKV) enforces extreme sparsity by firing exclusively on dynamic motion. We resolve dense topological and temporal SNN bottlenecks via Lateral Spiking Topology Routing (LSTR) for zero-MAC spatial propagation, and a Spiking State-Space (S3) Engine for robust long-range memory. Ultimately, S3T-Former establishes a new neuromorphic state-of-the-art—outperforming Spiking GCNs by up to +6.38\%—while consuming $<10\%$ the energy of comparable ANNs. Future work will explore adaptive localized firing to balance the sparsity-resolution trade-off in micro-motion recognition.
\bibliographystyle{ACM-Reference-Format}
\bibliography{sample-base}
\clearpage
\appendix
\begin{appendices}
In this supplementary material, we provide comprehensive additional details and analyses that were omitted from the main manuscript due to space constraints. Specifically, Appendix \ref{sec:app_datasets} elaborates on the specifics of the three large-scale skeleton datasets and their corresponding evaluation protocols. Following this, Appendix \ref{sec:app_implementation} outlines the detailed hyperparameter settings and training configurations to ensure the full reproducibility of our experiments. Moreover, Appendix \ref{sec:app_energy_sparsity} presents a rigorous theoretical analysis of the energy efficiency and Synaptic Operations (SOPs) of our S3T-Former, quantitatively proving its ultra-low power consumption. Finally, Appendix \ref{sec:app_visuals} offers extended qualitative visualizations, including additional topological routing maps, to further elucidate the internal reasoning mechanisms of our model.

\section{Datasets and Evaluation Protocols}
\label{sec:app_datasets}
\textbf{NTU RGB+D 60} \cite{shahroudy2016ntu} is a widely used large-scale benchmark for skeleton-based action recognition, comprising 56,880 video samples across 60 action classes. The actions are performed by 40 distinct subjects and captured simultaneously by three Microsoft Kinect v2 cameras from different viewing angles. Following the authoritative guidelines, we adopt two standard evaluation protocols: \textbf{Cross-Subject (X-Sub)}, where the 40 subjects are equally split into training and testing groups, and \textbf{Cross-View (X-View)}, where sequences captured by cameras 2 and 3 are utilized for training, while those from camera 1 are reserved for testing.

\textbf{NTU RGB+D 120} \cite{liu2019ntu} is the comprehensive extension of NTU RGB+D 60, currently serving as the most challenging and largest dataset in this domain. It encompasses 114,480 skeletal sequences spanning 120 action classes, performed by 106 subjects across 32 distinct camera setups. We strictly adhere to the official evaluation protocols: \textbf{Cross-Subject (X-Sub)}, which divides the 106 subjects into 53 training and 53 testing groups, and \textbf{Cross-Setup (X-Set)}, where the 32 camera setups are partitioned into even IDs for training and odd IDs for testing.

\textbf{Northwestern-UCLA (NW-UCLA)} \cite{wang2014cross} contains 1,494 video sequences representing 10 daily action categories. These actions are performed by 10 different subjects and captured by three Kinect cameras from varying viewpoints. Following the standard Cross-View evaluation protocol, we use samples from the first two cameras (V1 and V2) for training, and evaluate the model's performance on the samples from the third camera (V3).

\section{Implementation Details and Hyperparameters}
\label{sec:app_implementation}
\subsection{Implementation Details}

Our proposed S3T-Former is implemented using the PyTorch framework in conjunction with the open-source SpikingJelly library. To ensure full reproducibility, we detail the implementation configurations across three primary dimensions: network instantiation, data processing, and optimization strategy.

\textbf{Network Configurations and Spiking Dynamics.} 
The core backbone is constructed by stacking $L=6$ S3T Blocks, with the spatial topology routing decoupled into $H=8$ independent attention heads. The embedding dimension $D$ and the simulation time step $T$ are dynamically scaled to instantiate different model variants (e.g., $D \in \{256, 384\}$ and $T \in \{8, 16, 32\}$) for comprehensive comparisons. For the neuromorphic dynamics, we employ LIF neurons with a fixed firing threshold of $U_{th} = 0.5$ and a membrane leakage factor of $\tau = 0.5$. Since the binary spike generation (Heaviside step function) is inherently non-differentiable, we utilize the Arctangent (ATan) surrogate gradient function with a shape parameter of $\alpha = 2.0$ to enable stable backpropagation through time (BPTT).

\textbf{Data Preprocessing and Augmentation.}
To strictly align the raw skeleton data with the SNN simulation time windows, each action sequence is uniformly sampled or padded to a fixed temporal length $T$. The 3D spatial coordinates are zero-centered based on the root joint (the ``spine" node) to ensure translation invariance. Notably, we strictly refrain from employing any spatial or temporal skeleton data augmentation techniques (such as random rotation, scaling, or shear) during training. This deliberate constraint is designed to rigorously validate the inherent representational robustness and kinematic reasoning capabilities of our pure-spiking design, free from the performance inflation typically induced by heavy augmentations.

\textbf{Optimization and Temporal Efficient Training.}
The network is trained from scratch for 250 epochs using a total batch size of 64. We optimize the model using the AdamW optimizer with a base learning rate of 0.01 and a weight decay of 0.0005. To stabilize the initial learning phase, we apply a 10-epoch linear warmup, followed by a cosine annealing learning rate schedule that gradually decays to $1 \times 10^{-5}$. To effectively optimize the deep temporal dynamics without suffering from gradient vanishing, we apply the Temporal Efficient Training (TET) loss directly on the continuous outputs of our U-Readout integrator. Specifically, the cross-entropy loss is computed and averaged across all temporal simulation steps to provide dense objective supervision. All experiments are accelerated using Automatic Mixed Precision (AMP) to reduce memory footprint, seamlessly distributed across four NVIDIA Tesla V100 GPUs.
\lstset{
    backgroundcolor=\color{white},
    basicstyle=\fontsize{8pt}{9pt}\ttfamily\selectfont,
    keywordstyle=\color{blue}\bfseries,
    commentstyle=\color{green!50!black}\itshape,
    stringstyle=\color{red},
    showstringspaces=false,
    breaklines=true,
    frame=single,
    rulecolor=\color{gray!40},
    captionpos=b,
}

\section{PyTorch-like Pseudocode of S3T-Former}
\label{sec:app_pseudocode}

To facilitate future research and ensure strict reproducibility, we provide the PyTorch-style pseudocode for the core S3T-Attn in Listing 1. The code explicitly demonstrates that our architecture strictly operates on discrete binary spikes. Specifically, the local binding is executed via negligible bitwise AND operations (`*`), the spatial routing utilizes parameter-free conditional additions (`einsum` with binary operands), and the temporal integration is achieved through a linear-time $O(T)$ state-space loop. This implementation completely circumvents the dense $O(N^2 \times D)$ floating-point attention matrices typical of vanilla Transformers.

\lstset{
    backgroundcolor=\color{white},
    basicstyle=\fontsize{7.5pt}{8.5pt}\ttfamily\selectfont,
    keywordstyle=\color{blue}\bfseries,
    commentstyle=\color{green!50!black}\itshape,
    stringstyle=\color{red},
    showstringspaces=false,
    breaklines=true,
    frame=lines, 
    rulecolor=\color{gray!60},
    xleftmargin=1em,
    aboveskip=1em,
    belowskip=1em,
    captionpos=b,
}

\begin{lstlisting}[language=Python, caption=PyTorch-like Pseudocode of S3T-Former Attention.]
import torch
from torch import einsum

def s3t_attention(x, base_topo, learned_topo, alpha=0.8):
    # x: [T, B, C, N] (Time, Batch, Channels, Nodes)
    T, B, C, N = x.shape
    
    # 1. ATG-QKV Generation
    x_grad = torch.cat([torch.zeros_like(x[0:1]), torch.abs(x[1:]-x[:-1])])
    qkv_in = alpha * x_grad + (1 - alpha) * x  # Soft-ATG blending
    
    q = LIF_Q(Linear_Q(qkv_in)) # Binary motion queries {0, 1}
    k = LIF_K(Linear_K(qkv_in)) # Binary motion keys {0, 1}
    v = LIF_V(Linear_V(x))      # Binary static values {0, 1}
    
    # 2. Local Binding (Zero MACs, pure bitwise AND)
    kv_local = k * v 
    
    # 3. Lateral Spiking Topology Routing (LSTR)
    topo = torch.softmax(base_topo + learned_topo, dim=-1)
    # Conditional additions via binary operands
    kv_spatial = einsum('h i j, t b h j d -> t b h i d', topo, kv_local)
    kv_spike = LIF_Buffer(kv_spatial) 
    
    # 4. Spiking State-Space (S3) Engine
    decay = torch.clamp(torch.sigmoid(decay_w), 0.01, 0.99)
    M = torch.zeros_like(kv_spike[0])
    out = torch.empty_like(q)
    
    for t in range(T):
        M = decay * M + (1 - decay) * kv_spike[t] # Temporal accumulation
        out[t] = q[t] * M                         # Spike-gated retrieval
        
    # 5. Output Projection
    return LIF_Out(Linear_Out(out)) + x
\end{lstlisting}
\section{Hardware-Aware Energy and Spiking Sparsity Analysis}
\label{sec:app_energy_sparsity}

To quantitatively validate the low-power characteristics of the proposed S3T-Former, we conduct a hardware-aware energy consumption analysis based on the standard 45nm CMOS technology \cite{horowitz20141}. In standard Artificial Neural Networks (ANNs), the primary computational cost stems from dense floating-point Multiply-Accumulate (MAC) operations. The theoretical energy consumption for an ANN is directly proportional to its total FLOPs ($E_{\text{ANN}} = \mathcal{O}_{\text{FLOPs}} \times E_{\text{MAC}}$), where $E_{\text{MAC}} = 4.6 \text{ pJ}$ represents the energy cost of a 32-bit MAC operation.

For our S3T-Former, the computational graph is fundamentally hybrid. Based on the discrete or continuous nature of the input tensors, the overall operations are strictly decoupled into floating-point MACs and spike-driven Synaptic Operations (SOPs). Furthermore, during inference, all Batch Normalization (BN) operations are analytically folded into their preceding linear layers, incurring zero additional computational energy.

\textbf{1. Floating-Point MAC Operations:} 
MAC operations exclusively occur when processing continuous values, including the representation extraction in M-ASE, temporal integration within the S3-Engine, and the classification head. The total MAC operations are the sum of their respective FLOPs:
\begin{equation}
    \mathcal{O}_{\text{MAC}} = \mathcal{O}_{\text{ASE}} + \sum_{l=1}^{L} (\mathcal{O}_{\text{S3-Engine}}^{(l)} + \mathcal{O}_{\text{proj}}^{(l)}) + \mathcal{O}_{\text{FC\_head}}.
\end{equation}

\textbf{2. Spike-Driven Synaptic Operations (SOPs):}
The core efficiency originates from the spike-driven modules, including the ATG-QKV generation, the Lateral Spiking Topology Routing (LSTR), and the Spiking FFN (MLP). Because the inputs to these layers are binary spikes $S \in \{0, 1\}$, dense matrix multiplications mathematically degenerate into sparse Accumulate (AC) operations ($E_{\text{AC}} = 0.9 \text{ pJ}$). The total number of SOPs is intrinsically tied to the empirical spike firing rate ($Fr$):
\begin{equation}
    \mathcal{O}_{\text{SOP}} = \sum_{l=1}^{L} \sum_{m \in \mathcal{M}_{\text{spike}}} \mathcal{O}_{m}^{(l)} \times Fr_{m}^{(l)},
\end{equation}
where $\mathcal{M}_{\text{spike}} = \{Q, K, V, \text{LSTR}, \text{MLP1}, \text{MLP2}\}$. Notably, local binding ($K \odot V$) reduces to bitwise AND operations, the energy cost of which is negligible compared to ACs. 

\textbf{Empirical Sparsity and Energy Estimation:}
To empirically substantiate the SOP reduction, we systematically profile the layer-wise spike firing rates ($Fr$) of our standard S3T-Former ($T=16$). As reported in Table \ref{tab:firing_rate}, thanks to the dynamic filtering of the LSTR module and the event-driven nature of our architecture, the firing rates of the critical bottlenecks plummet as the network deepens. For instance, the Topo Buffer firing rate drops from 10.09\% in Block 1 to a mere 1.91\% in Block 6. Similarly, the MLP 1 layer—which scales the channel dimension by $4\times$ and encompasses the majority of the network's parameters—maintains an extreme sparsity ranging from 4.20\% down to 1.38\%. 

Given that an AC operation consumes roughly $5\times$ less energy than a MAC operation, the theoretical energy for the spike-driven modules is strictly bounded by $\sim \frac{1}{5} \times Fr$. With our global average firing rate at $\sim 16.9\%$ and the most parameter-heavy expansion layers operating at $<3\%$ sparsity, the SOP energy is drastically minimized. Even when accounting for the continuous MAC operations in the M-ASE and S3-Engine, the total theoretical energy consumption ($E_{\text{SNN}} = \mathcal{O}_{\text{MAC}} \times E_{\text{MAC}} + \mathcal{O}_{\text{SOP}} \times E_{\text{AC}}$) is conservatively estimated to be strictly within \textbf{10\%} of a structurally equivalent dense ANN. This rigorously confirms the absolute power superiority of our neuromorphic design without compromising recognition accuracy.

\begin{table*}[t]
\caption{Layer-wise average spike firing rates ($Fr$, \%) of the standard S3T-Former ($T=16$). MLP 1 and MLP 2 denote the channel expansion ($4\times$) and projection layers within the Spiking FFN, respectively. The most parameter-heavy expansion modules (Topo Buffer and MLP 1) exhibit extreme deep-layer sparsity.}
\label{tab:firing_rate}
\centering
\setlength{\tabcolsep}{5pt}
\begin{tabular}{c | c c c | c c | c c}
\toprule
\multirow{2}{*}{\textbf{Block}} & \multicolumn{3}{c|}{\textbf{ATG-QKV Generation}} & \multicolumn{2}{c|}{\textbf{Spatial-Temporal Routing(LSTR+S3)}} & \multicolumn{2}{c}{\textbf{Spiking FFN (MLP)}} \\
\cmidrule{2-8}
 & $Q$ & $K$ & $V$ & Topo Buffer & Attn Out & MLP 1 (Exp.) & MLP 2 (Proj.) \\
\midrule
1 & 27.49 & 53.00 & 44.24 & 10.09 & 28.15 & 4.20 & 20.60 \\
2 & 21.43 & 49.31 & 27.71 & 2.95  & 23.17 & 2.45 & 17.62 \\
3 & 24.77 & 56.07 & 24.19 & 1.86  & 25.90 & 1.90 & 20.53 \\
4 & 24.26 & 53.65 & 22.42 & 1.32  & 22.93 & 1.56 & 28.19 \\
5 & 19.77 & 55.84 & 19.13 & 1.31  & 30.01 & 1.38 & 34.24 \\
6 & 16.30 & 67.10 & 19.65 & 1.91  & 41.76 & 1.54 & 45.23 \\
\midrule
\rowcolor[gray]{0.95}
\textbf{Avg.} & \textbf{22.33} & \textbf{55.82} & \textbf{26.22} & \textbf{3.24} & \textbf{28.65} & \textbf{2.17} & \textbf{27.73} \\
\bottomrule
\end{tabular}
\end{table*}
\section{Extended Qualitative Results}
\label{sec:app_visuals}
\subsection{Full Confusion Matrix and Failure Diagnosis}
\begin{figure}[h]
    \centering
    \includegraphics[width=0.5\textwidth]{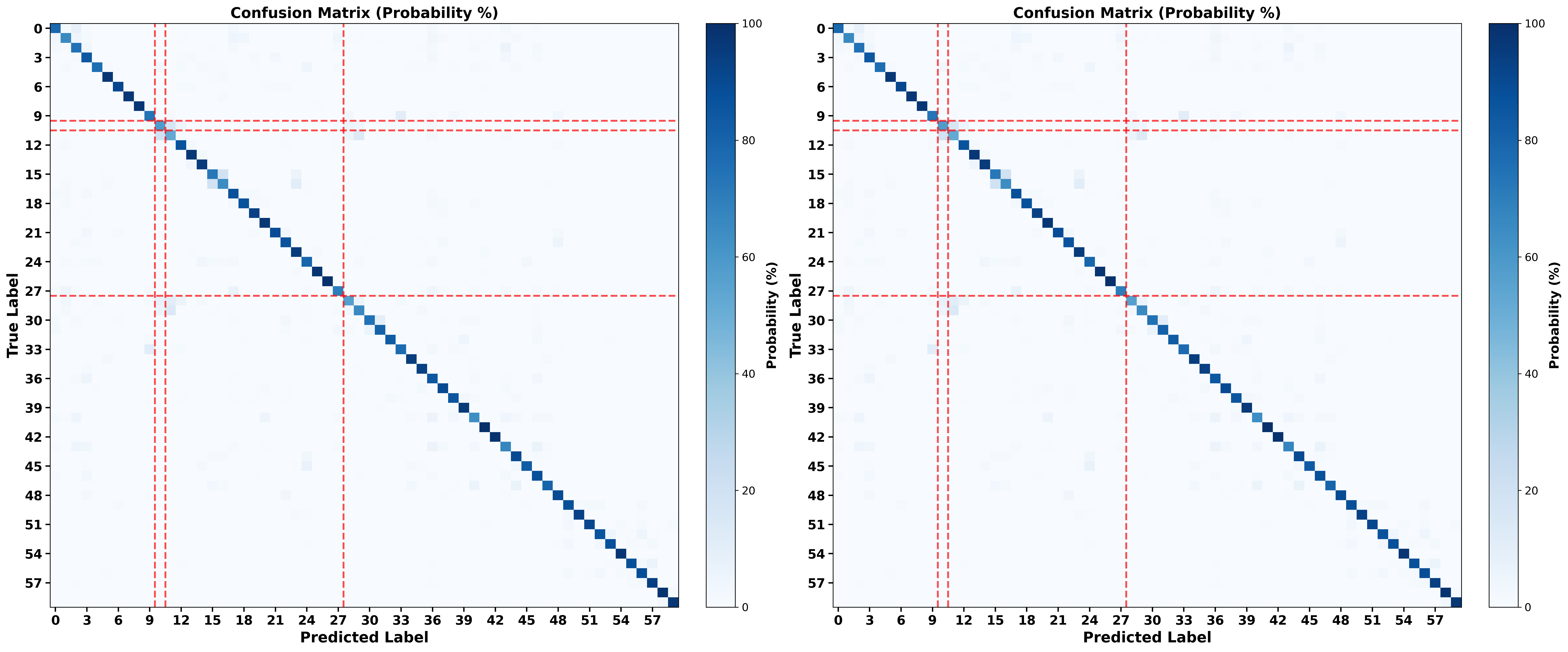} 
    \caption{Full confusion matrix of S3T-Former on the NTU RGB+D 60 benchmark. The red dashed lines explicitly highlight the top three most confusing action classes: Class 10 (Reading), Class 11 (Writing), and Class 28 (Playing with phone/tablet). These failures highlight the inherent difficulty of distinguishing highly localized micro-motions within an aggressively sparse, event-driven SNN framework.}
    \label{fig:app_confusion}
\end{figure}

For absolute transparency and to provide deeper insights into the boundary conditions of our neuromorphic design, we present the full confusion matrix for the NTU RGB+D 60 evaluation in Fig. \ref{fig:app_confusion}. 

While S3T-Former achieves a strong overall accuracy of 84.94\%, a diagnostic inspection of the confusion matrix reveals a highly specific and consistent failure mode. The red dashed lines in Fig. \ref{fig:app_confusion} highlight the top three most challenging categories: Class 11 (Error Rate: 47.79\%), Class 28 (Error Rate: 43.64\%), and Class 10 (Error Rate: 41.76\%). In the NTU RGB+D 60 dataset, these classes correspond to "Writing," "Playing with phone/tablet," and "Reading," respectively.

These actions share a fundamental biomechanical similarity: they involve seated postures with extremely fine-grained, localized micro-motions of the fingers and wrists, accompanied by minimal macro-skeletal displacement. Because our Asymmetric Temporal-Gradient QKV (ATG-QKV) mechanism is explicitly engineered to aggressively filter out static spatial features and fire exclusively on physical motion gradients to maximize energy efficiency, it generates an exceptionally sparse event stream for these specific actions. Consequently, the network inadvertently sacrifices the high-resolution static spatial cues necessary to robustly distinguish between identical hand-to-chest proximities (e.g., holding a book versus holding a phone). 

This diagnostic reveals an inherent trade-off in current pure-spiking architectures between extreme temporal sparsity and micro-motion spatial resolution. Addressing this bottleneck—potentially through multi-scale dynamic thresholds or localized firing amplification—remains an exciting avenue for advancing neuromorphic action recognition.

\subsection{Layer-wise Skeleton Spiking Heatmaps: The Semantic Sparsity Funnel}
\label{sec:app_semantic_funnel}

\begin{figure*}[htbp] 
    \centering
    \begin{subfigure}[b]{0.95\textwidth}
        \centering
        \includegraphics[width=\textwidth]{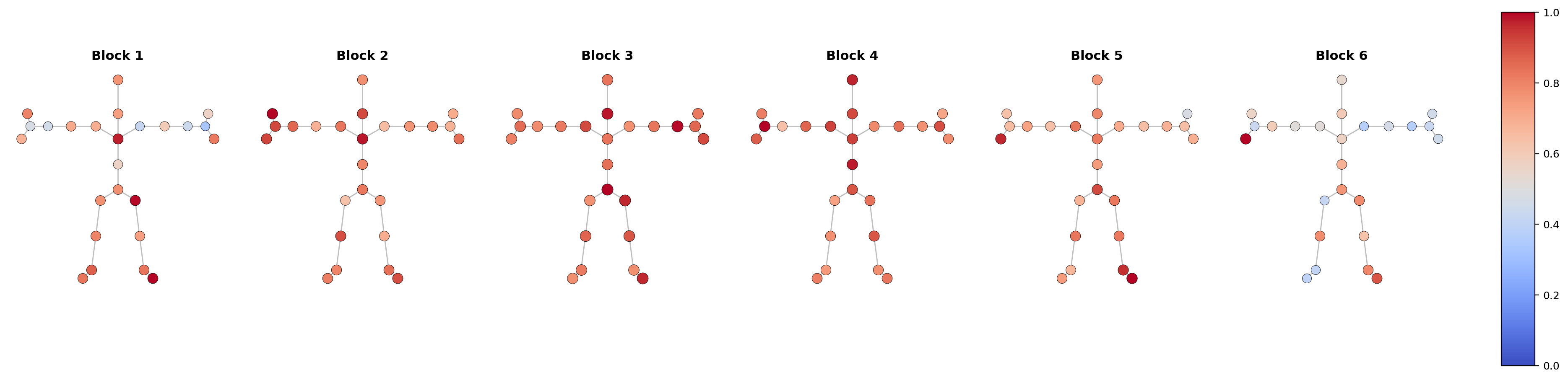}
        \caption{Class 6: Throwing}
    \end{subfigure}
    \begin{subfigure}[b]{0.95\textwidth}
        \centering
        \includegraphics[width=\textwidth]{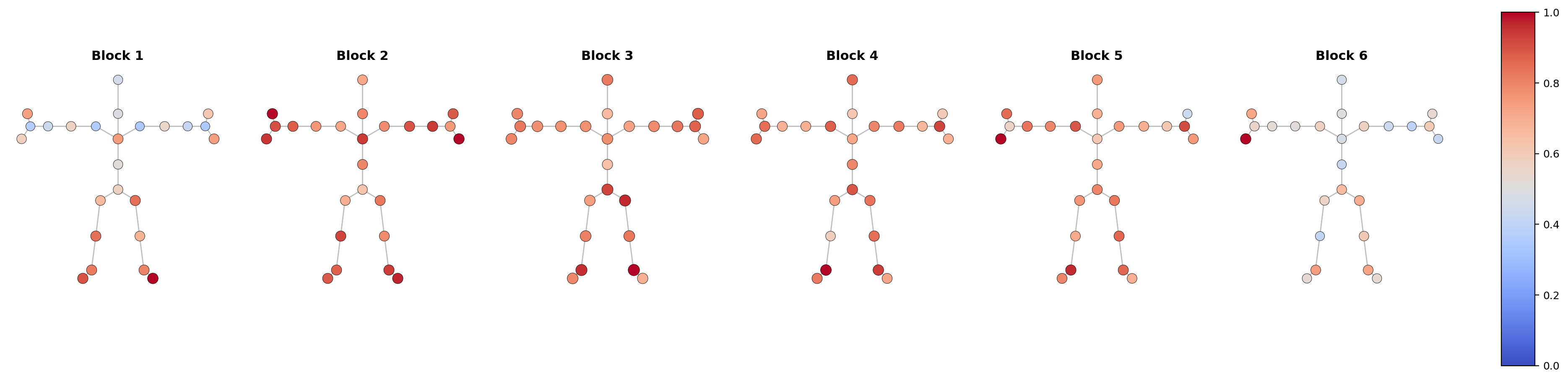}
        \caption{Class 23: Kicking something}
    \end{subfigure}
    \begin{subfigure}[b]{0.95\textwidth}
        \centering
        \includegraphics[width=\textwidth]{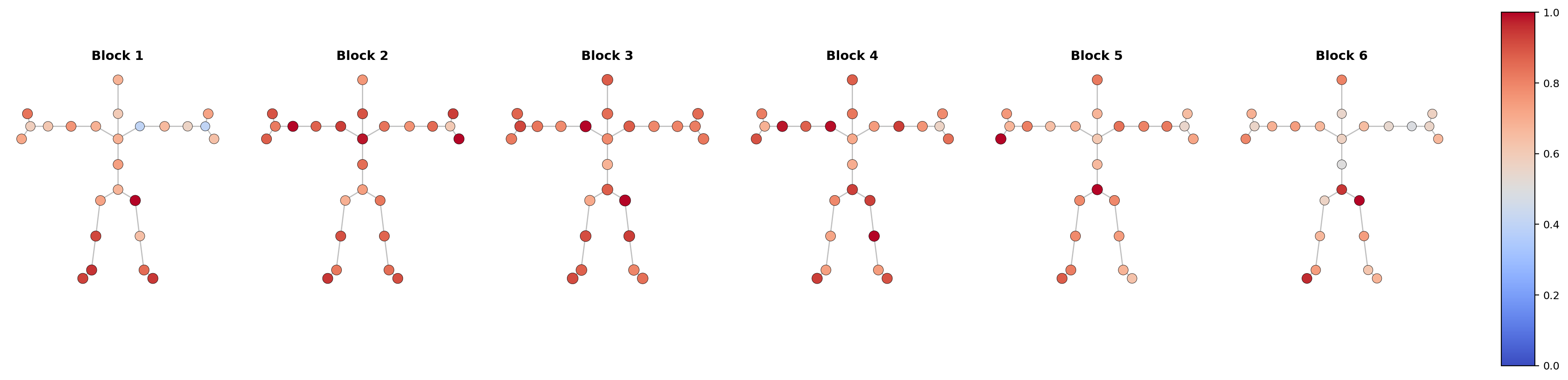}
        \caption{Class 26: Jump up}
    \end{subfigure}
    \begin{subfigure}[b]{0.95\textwidth}
        \centering
        \includegraphics[width=\textwidth]{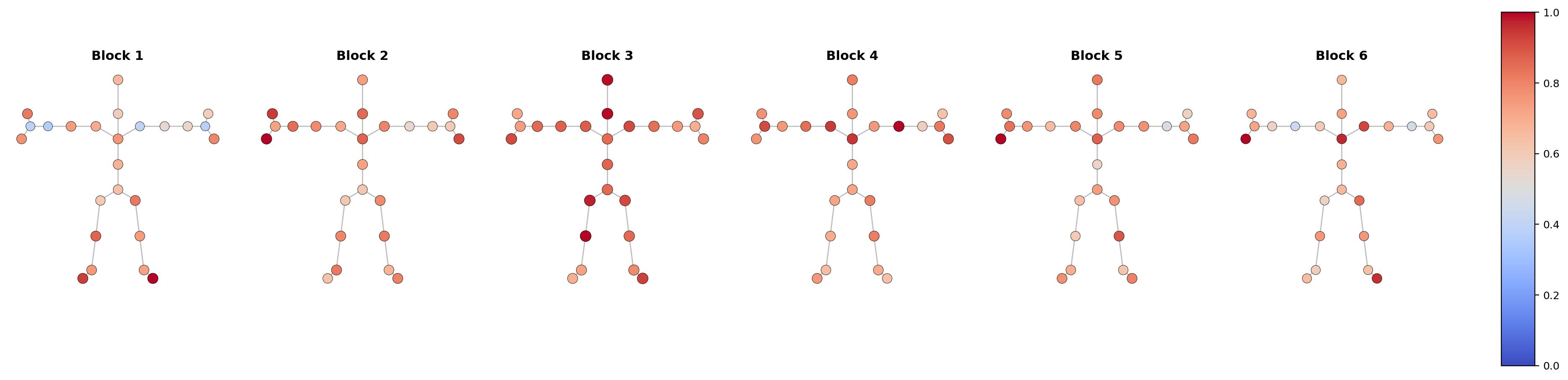}
        \caption{Class 42: Falling}
    \end{subfigure}
    \begin{subfigure}[b]{0.95\textwidth}
        \centering
        \includegraphics[width=\textwidth]{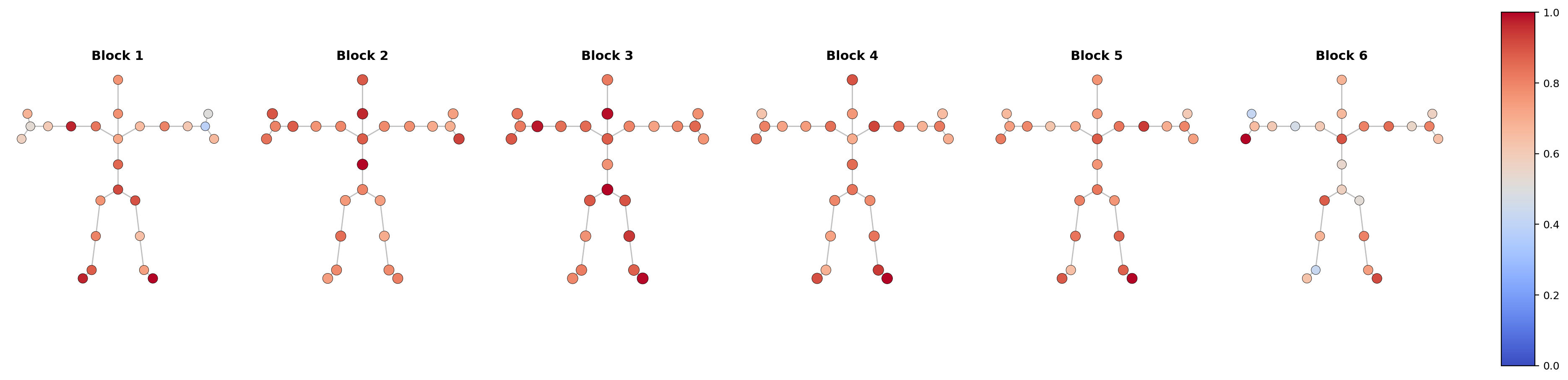}
        \caption{Class 49: Punch/Slap other person}
    \end{subfigure}
    \caption{Layer-wise evolution of spiking heatmaps. Red indicates active firing, while blue denotes zero-energy dormancy. This illustrates how the network funnels dense low-level kinematics into ultra-sparse semantic endpoints at deeper layers.}
    \label{fig:app_funnel}
\end{figure*}

To intuitively visualize the extreme spatial sparsity and the layer-wise representational shift achieved by our architecture, we project the accumulated query ($Q$) spike firing rates directly onto the 3D human skeleton. In Fig. \ref{fig:app_funnel}, we present the layer-wise evolution (from Block 1 to Block 6) across five distinct action categories: Throwing (Class 6), Kicking (Class 23), Jumping Up (Class 26), Falling (Class 42), and Punch/Slap (Class 49). 

The visualizations reveal a profound "sparsity funnel" effect, demonstrating that our S3T-Former does not merely perform "on-demand energy allocation," but effectively acts as a powerful \textbf{Spatio-Temporal Semantic Hard-Attention} mechanism. The progression follows three distinct representational phases:

\begin{itemize}
    \item \textbf{Kinematic Tracking Phase (Blocks 1-2):} In the shallow layers, the network faithfully tracks the raw physical kinetic chain. For instance, in the "Kicking" (Class 23) and "Throwing" (Class 6) actions, the entire active limb (leg or arm) alongside the stabilizing torso exhibits high firing rates (red). The network allocates energy to capture the low-level physical deformations across multiple related joints.
    \item \textbf{Spatial Pruning Phase (Blocks 3-4):} As representations propagate deeper, the Lateral Spiking Topology Routing (LSTR) actively filters out redundant spatial noise. Peripheral joints that only serve as physical connectors begin to go dormant (turning light blue), while the core action centers remain highly active.
    \item \textbf{Semantic Focus Phase (Blocks 5-6):} In the deepest layers, the network achieves extreme sparsity. The physical skeleton becomes largely dormant, and the remaining spikes are exclusively clustered at the most discriminative topological endpoints defining the action's semantic core. For striking actions like "Punching" (Class 49) and "Kicking" (Class 23), the energy is sharply focused on the extreme end-effectors (the fist and the ankle). Fascinatingly, for global body momentum actions like "Falling" (Class 42) and "Jumping" (Class 26), the deep layers shift their focus toward the spine and the center of gravity. 
\end{itemize}

This visual evidence mathematically and biologically validates our design: as the feature dimensions expand in deeper stages, the network systematically zeroes out irrelevant physical tracking, condensing the complex input into ultra-sparse, highly discriminative semantic signatures. This guarantees our minimal energy footprint while simultaneously boosting recognition accuracy.
\subsubsection{Per-Class Performance Breakthrough}
\label{sec:Per-Class Performance Breakthrough}
\begin{figure}[t]
    \centering
    \includegraphics[width=0.45\textwidth]{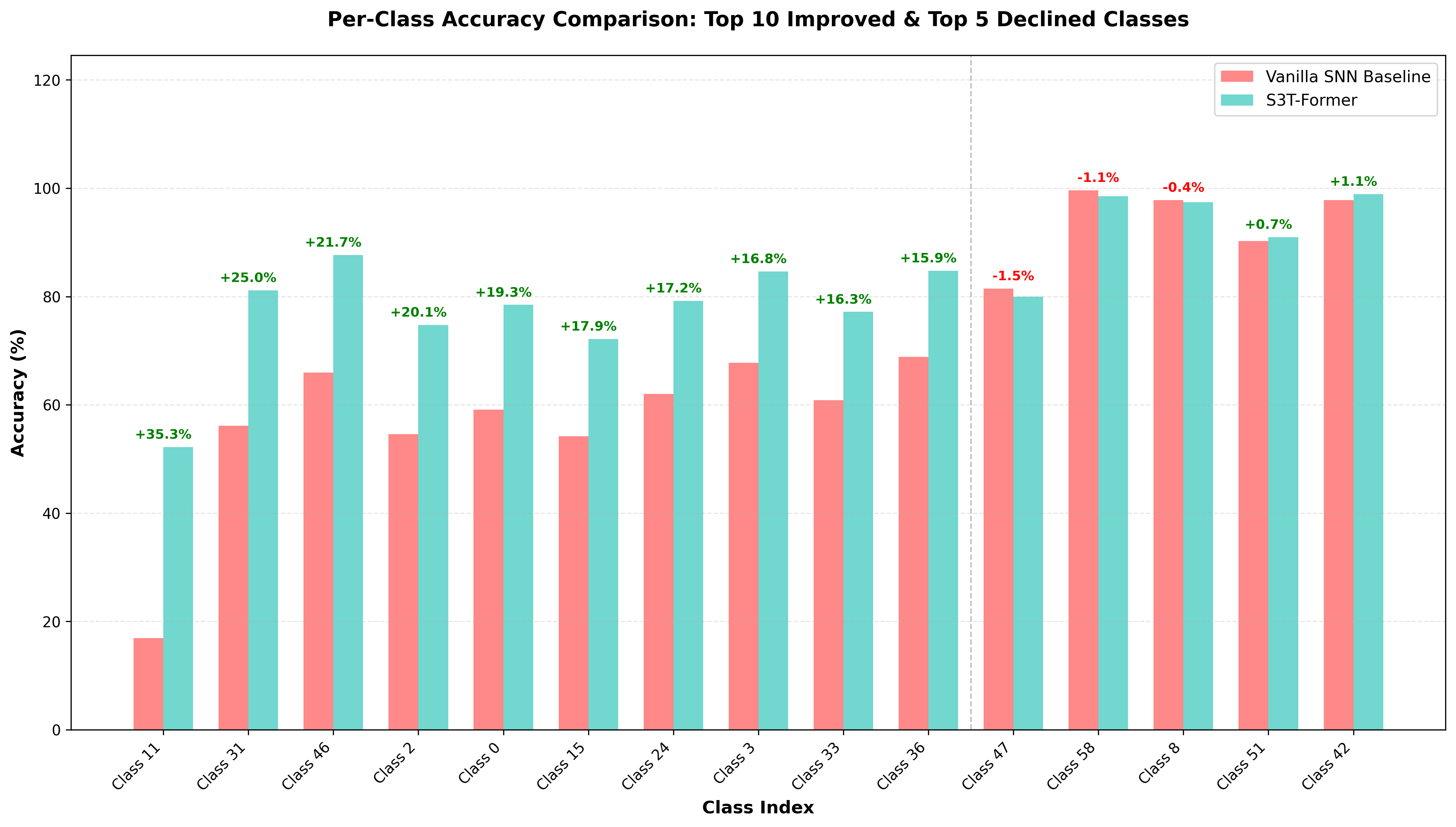} 
    \caption{Per-class Top-1 accuracy (\%) comparison between the Vanilla SNN baseline and our S3T-Former on the NTU RGB+D 60 dataset. The left section highlights the Top-10 classes with the most significant absolute accuracy improvements, while the right section shows the Top-5 classes with slight declines or minimal gains. S3T-Former demonstrates overwhelming superiority on complex, fine-grained actions while maintaining extreme robustness on saturated macro-motions.}
    \label{fig:per_class_comparison}
\end{figure}

To granularly evaluate the robustness of S3T-Former and diagnose its specific advantages over traditional neuromorphic architectures, we visualize a per-class accuracy comparison against the Vanilla SNN baseline in Fig. \ref{fig:per_class_comparison}. The results reveal a profound paradigm shift in handling highly complex and fine-grained kinematics.
\textbf{Massive Gains on Fine-Grained Kinematics.} The left sector of the chart highlights an overwhelming superiority of S3T-Former on actions requiring precise temporal tracking and localized micro-motion perception. Most notably, for ``Writing'' (Class 11), the baseline SNN almost completely fails (achieving merely $\sim$17\% accuracy) due to severe short-term amnesia and the inability to distinguish subtle hand-joint displacements from static noise. In stark contrast, S3T-Former yields an astounding absolute improvement of +35.3\%. Similar leap-forward gains are observed in other localized or self-occluded actions, such as ``Taking a selfie'' (Class 31, +25.0\%), ``Touch neck'' (Class 46, +21.7\%), and ``Brushing teeth'' (Class 2, +20.1\%). This compelling evidence proves that our Multi-Stream Anatomical Spiking Embedding (M-ASE) and the Spiking State-Space (S3) Engine successfully empower the network to capture multi-order motion gradients and long-term causalities that traditional LIF neurons inherently lose.

\textbf{Robustness on Saturated Macro-Motions.} The right sector displays classes where our model exhibits marginal declines (e.g., -1.5\% for Class 47 ``Nausea/vomiting'', -1.1\% for Class 58 ``Walking towards each other'') or minimal gains (e.g., Class 42 ``Falling''). Crucially, the baseline model already achieves highly saturated performances ($>80\%$ to nearly $100\%$) on these specific actions, as they involve unmistakable, large-scale macro-displacements that are easily distinguishable even for primitive networks. The trivial fluctuations ($<1.5\%$) on these saturated classes confirm that our ATG-QKV despite aggressively pruning static spatial topology to enforce extreme energy sparsity—does not degrade the global structural representation. Ultimately, S3T-Former achieves massive accuracy breakthroughs on the hardest categories without sacrificing the fundamental stability of the easiest ones.
\end{appendices}

\end{document}